\documentclass{article}
\usepackage{iclr2027_conference,times}

\usepackage{graphicx}
\usepackage{booktabs}
\usepackage{amsmath}
\usepackage{amssymb}
\usepackage{multirow}
\usepackage{xcolor}
\usepackage{xspace}
\usepackage[normalem]{ulem}
\usepackage{wrapfig}
\usepackage{float}
\usepackage{flafter}
\usepackage{hyperref}
\usepackage{url}

\usepackage{amsmath,amsfonts,bm}

\def\eqref#1{equation~\ref{#1}}

\def\1{\bm{1}}

\DeclareMathAlphabet{\mathsfit}{\encodingdefault}{\sfdefault}{m}{sl}
\SetMathAlphabet{\mathsfit}{bold}{\encodingdefault}{\sfdefault}{bx}{n}

\newcommand{\eg}{\emph{e.g.}\xspace}
\newcommand{\ie}{\emph{i.e.}\xspace}

\newcommand{\scriptremove}[1]{\textcolor{red}{\sout{#1}}}
\newcommand{\scriptadd}[1]{\textcolor{green!50!black}{#1}}

\title{COMIC: Agentic Sketch Comedy Generation}

\author{Susung Hong \quad Brian Curless \quad Ira Kemelmacher-Shlizerman \quad Steve Seitz \\
University of Washington}

\iclrfinalcopy

\begin{document}

\maketitle
\ificlrfinal
  \lhead{}
\fi

\begin{abstract}

We propose a fully automated AI system that produces short comedic videos similar to sketch shows such as Saturday Night Live. Starting from character references, the system employs a population of agents loosely modeled on roles in real production studios and structured to optimize the quality and diversity of ideas and outputs through iterative competition, evaluation, and refinement. A key contribution is the introduction of LLM-based critics aligned with real viewer preferences through the analysis of a corpus of comedy videos on YouTube, enabling automatic evaluation of humor. We further propose multi-island relativistic evolution for script improvement, along with script-conditioned rendering critics that perform shot- and video-level tournaments. In both human and automated evaluations, COMIC receives higher scores than agentic and video generation baselines.

\end{abstract}

\vspace{-15pt}
\begin{figure}[H]
  \centering
  \includegraphics[width=\linewidth]{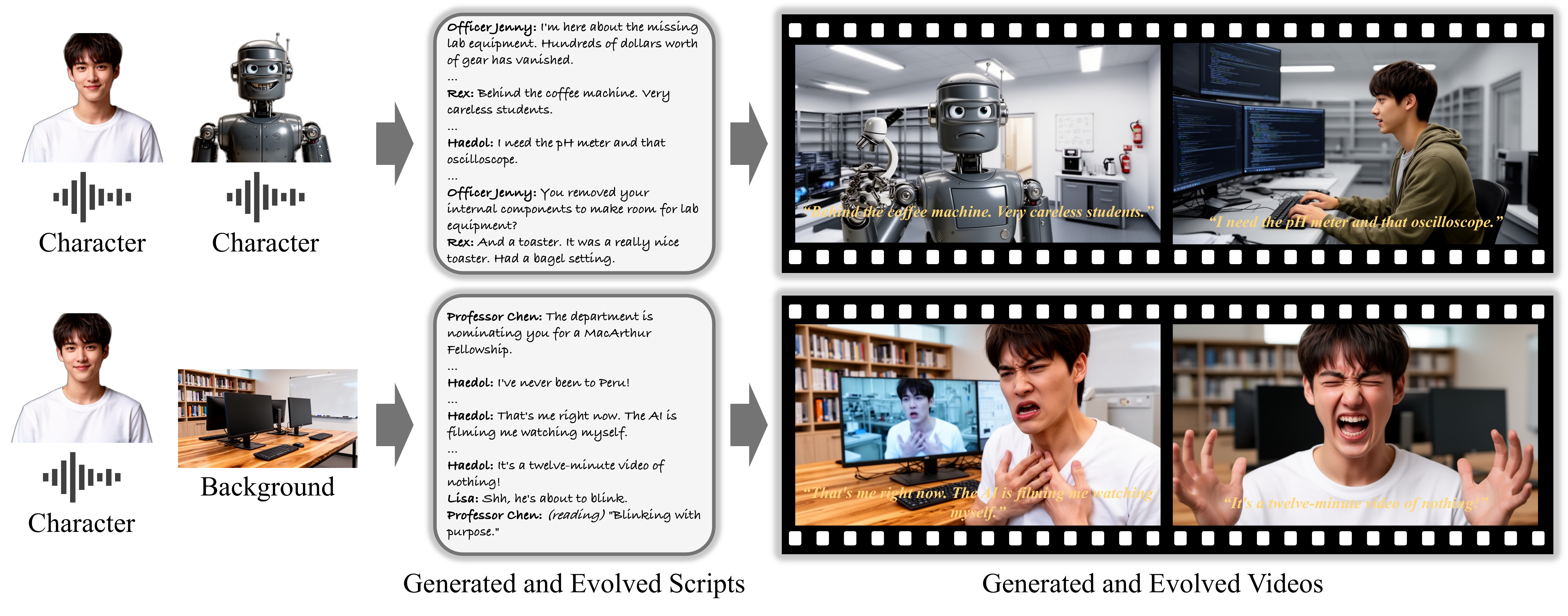}
  \vspace{-20pt}
  \caption{\textbf{COMIC} takes images, voices, and brief descriptions as input, and automatically
  generates entertaining comedy scripts along with corresponding video and audio.
  Our framework flexibly builds stories around multiple characters and custom
  backgrounds. Each generated comedy video is 1--2 minutes long; please watch them
  \ificlrfinal at \url{https://susunghong.github.io/COMIC}\else in the
  supplementary material\fi.}
  \label{fig:teaser}
\end{figure}
\vspace{-15pt}

\section{Introduction}

Can AI be funny? If you ask today's AI models to tell you a joke, you will likely receive a groan-inducing pun or a ``dad joke.'' While generative models now excel across a wide range of writing, coding, and media generation tasks, humor remains particularly challenging. This is not to say that LLMs are incapable of humor---with the right prompts and enough iterations, one can find some gems. However, reliably producing content that can make an audience laugh is difficult.

In this paper, we propose a fully automated framework, \textbf{COMIC} (\textbf{C}ontent \textbf{O}ptimization via \textbf{M}ulti-agent \textbf{I}terative \textbf{C}ompetition), for generating short comedic videos. The input to the system is a list of character descriptions (text, image, voice) and/or background references (Figure~\ref{fig:teaser}). Achieving this goal requires solving three distinct tasks automatically: conceptualizing the right comedic scenario, producing a funny script, and generating a high-quality, consistent, and engaging video. Each of these tasks is challenging. Crafting genuinely funny ideas and scripts requires navigating the subjective, multidimensional space of humor, and producing long-form video remains an open problem, as state-of-the-art video models typically produce only short clips and lack strong controls for inter-clip consistency.

Our approach starts from the observation that LLMs do occasionally produce humorous content when provided with the right structure. It is a bit like panning for gold---one must dig deep enough to gather sufficient material, then sift through it to find the humorous nuggets from which to build a sketch. This mirrors how real sketch comedy shows operate~\citep{marx2016live}, with groups of writers spending many hours brainstorming and iterating before converging on a set of finalists. However, evaluating humor automatically is a big challenge. We therefore derive LLM-based {\em writing critics} aligned with human preferences by analyzing a corpus of YouTube comedy sketch videos and their associated viewer engagement measured by projected views.

We build a comprehensive multi-agent system loosely modeled on human production studios, using agentic versions of roles such as scriptwriters, editors, and directors, with a structure designed to encourage diversity of ideas and optimize for the emergence of humorous content. Specifically, we instantiate multiple distinct {\em islands} of scripts, each governed by a writing-critic committee representing a different philosophy or niche. Script populations improve on each island through round-robin tournaments in which losing scripts are refined using feedback from the winners. This topology captures the multimodal nature of humor: ``good'' comedy can be slapstick, dry, or surreal, and success can manifest through various approaches.

Once scripts are refined, scene director agents break each script down into distinct shots---each with its own setup, \eg, characters, dialogue, expressions, and backgrounds---and render videos for each shot. Shots are produced consecutively, allowing the scene director for the current shot to reference previous shots for continuity. Each shot is evaluated by a set of script-conditioned rendering critics that embody diverse interpretations of how the narrative should be visually realized, and then refined based on critic feedback. This iterative pipeline employs depth- and breadth-wise competitions over refinement histories and scene realizations, respectively.

COMIC targets the generation of comedic videos, an open-ended creative task at the opposite extreme from mathematics or coding, which have objectively correct answers. As a systems contribution, it brings the full production workflow into one modular pipeline, from concept development and scriptwriting to storyboarding, audio-visual synthesis, continuity management, and quality control. We further demonstrate our key algorithmic contributions: generating and selecting writing critics aligned with real viewer engagement, island-based relativistic evolution for open-ended script improvement, and script-conditioned rendering critics with iterative shot- and video-level tournaments. Our experiments, across both human evaluation and independently selected benchmark video critics, show higher scores for COMIC than for the agentic and video generation baselines.

\section{Background}
\paragraph{Multi-agent evolutionary systems.}
Evolutionary computation has been applied to creative domains through genetic algorithms~\citep{sims1991artificial}, quality-diversity methods~\citep{mouret2015illuminating}, and distributed island models~\citep{tanese1989distributed,cantu1998survey,whitley1999island,novikov2025alphaevolve}. Recent LLM work applies related ideas to prompt optimization~\citep{yang2023large,fernando2023promptbreeder}, heuristic discovery~\citep{liu2024evolution}, and agent improvement~\citep{zhang2025darwin,yuan2025evoagent}. COMIC brings this perspective to comedy, an extremely open-ended domain, through competition among multiple aligned critics.

\paragraph{Video generation.}
Text-to-video models such as Sora~\citep{openai2025sora}, Veo~\citep{google2025veo}, Movie Gen~\citep{meta2024moviegen}, and Wan~\citep{wan2025wan} generate increasingly realistic clips, while controllable systems add signals such as audio~\citep{lin2025omnihuman,hong2025musicinfuser}. Extensions such as StreamingT2V~\citep{henschel2025streamingt2v} and FramePack~\citep{zhang2025packing} increase duration, but do not address narrative coherence or comedic quality. COMIC builds a long-form production system on top of these short-form capabilities.

\paragraph{Agentic video production.}
LLM-based systems use agents for direction, planning, and self-improvement~\citep{hong2023direct2v,huang2023free,lin2023videodirectorgpt,lian2023llm,long2025vista}. Vlogger~\citep{zhuang2024vlogger} decomposes minute-level vlog production into script, actor, video, and voice stages, while Captain Cinema~\citep{xiao2026captain} combines top-down keyframe planning with long-context synthesis for multi-scene movies. Storyboard-based methods~\citep{zheng2024videogen,wu2025automated,shi2025animaker,huang2025filmaster,li2024anim,dalal2025one} also target longer videos, but use shallow, single-pass workflows. COMIC instead searches the creative space through iterative competition under diverse, aligned critics.

\begin{figure}[t]
\centering
\includegraphics[width=\linewidth]{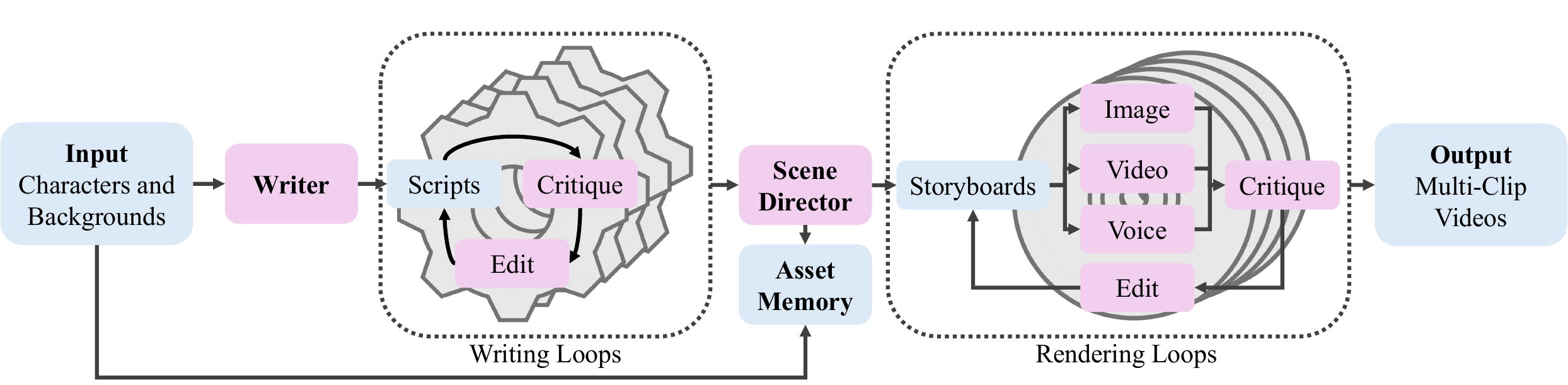}
\vspace{-20pt}
\caption{Overall agentic flow. COMIC is loosely modeled on a production studio, with agentic counterparts for a writer, critic, editor, and scene director. The writing and rendering loops generate scripts and videos through island-based competition and iteration.}
\vspace{-10pt}
\label{fig:overall}
\end{figure}

\section{Content Optimization via Multi-Agent Iterative Competition (COMIC)}
\label{sec:comic}

\subsection{Problem Statement}
\label{sec:problem}

We address \emph{automated sketch comedy video generation}: given character specifications $\mathcal{X} = [x_1, \ldots, x_C]$---each comprising a portrait image, voice sample, and text description---and background assets $\mathcal{B} = [b_1, \ldots, b_M]$, the system must produce a short video $\mathcal{V}^*$ that is narratively coherent, visually consistent, and funny to human viewers.

Figure~\ref{fig:overall} illustrates the two coupled subproblems. \emph{Script generation} synthesizes $s^* \in \mathcal{S}$ through writing, comparison, and revision. \emph{Visual realization} translates $s^*$ into a shot sequence $\mathcal{V}=[v_1,\ldots,v_N]$ using storyboards, multimodal generation, and rendering-critic refinement.

However, a single-pass instantiation is insufficient for high-quality results; good scripts and videos are forged through multiple rounds of revision. COMIC uses human-aligned critics (Section~\ref{sec:critic}), evolves scripts through competitive island-based search (Section~\ref{sec:writing}), and realizes them audio-visually through iterative rendering and critic refinement (Section~\ref{sec:rendering}).

Crucially, throughout the process, we sample and leverage distinct types of critics. \emph{Writing critics} compare and refine scripts during island evolution, while script-conditioned \emph{rendering critics} refine and select shots and videos. \emph{Benchmark video critics} are independently sampled and used solely for held-out automated evaluation (Section~\ref{sec:baseline_comparison}).

\subsection{Why Fixed Objectives Fall Short}
\label{sec:why_fail}

Traditional goal-based optimization presupposes a stationary reward function $R \colon \mathcal{S} \to \mathbb{R}$. However, humor is inherently context-dependent and subjective. A fixed scalar objective invites Goodhart's Law~\citep{goodhart1984problems}, rewarding a proxy metric rather than genuine creative quality. Furthermore, great humor takes many distinct forms; slapstick and dry wit share no common measuring stick, and different people prefer different styles. It is impossible to aggregate preference profiles into a single, consistent ranking without sacrificing desirable properties~\citep{arrow1950difficulty}. Meanwhile, existing agentic video systems~\citep{lin2023videodirectorgpt, zheng2024videogen, wu2025automated} rely on fixed roles in a shallow pipeline, limiting feedback and alternative perspectives. In contrast, COMIC introduces multiple relativistic islands, each with its own comparative objective, perspective, and niche. Within the framework, a script's fitness is evaluated relative to current competitors on the same island, and losing scripts are constantly refined using critic feedback. Consequently, the multidimensional competitive baseline can continuously rise without requiring a fixed destination.

\subsection{Alignment of Writing Critics to Real Viewers}
\label{sec:critic}

Writing quality depends critically on writing-critic selection. Rather than hand-crafting prompts or fine-tuning a dedicated evaluator, we use a \emph{generate-and-select} strategy based on a large corpus of videos. We synthesize a diverse pool of candidate writing critics, each defined by a system prompt specifying its persona, and retain those whose preferences align with empirical audience engagement signals. This provides diverse evaluative perspectives while enabling aggressive pruning to retain only informative critics.

\paragraph{Engagement scoring.}
We collect 4,940 data points from five YouTube sketch comedy channels: Foil Arms \& Hog, Key \& Peele, SNL, Studio C, and Viva La Dirt League. We use view counts as a proxy for popularity. Since view-count trajectories empirically follow an S-curve, we fit a per-channel logistic growth model, $V(t)=L/(1+\exp(-r(t-t_0)))$, by nonlinear least squares. We define each video's \emph{engagement score} as its projected infinite-horizon view count, $L_{\text{proj}}=V(t)\cdot(1+\exp(-r(t-t_0)))$. Scripts are then selected for top, middle, and bottom engagement tiers and partitioned into $\mathcal{S}_{\text{in-context}}$, $\mathcal{S}_{\text{val}}$, and held-out $\mathcal{S}_{\text{test}}$. The per-channel fitted trajectories are shown in Appendix~\ref{sec:appendix_critic_selection}, and additional processing details are provided there.

\paragraph{Writing-critic pool generation.}
We construct a diverse pool $\mathcal{C}_{\text{pool}}$ by prompting a meta-critic $p_\text{script}$ with stratified, tier-labeled examples:
\begin{equation}
\mathcal{C}_{\text{pool}} \sim p_{\text{script}}(\mathcal{C}\mid\mathcal{S}_{\text{in-context}}),
\end{equation}
This calibrates each writing critic's aesthetic preferences to channel-specific engagement patterns. Rather than producing a single critic, we sample a set of critics (10 in practice) and select the subset that best discriminates real viewer engagement scores.

\paragraph{Task-specific selection.}
To capture both coarse and fine quality distinctions, we define two pairwise tasks. \emph{Top vs.\ Bottom} targets critics sensitive to large quality gaps, while \emph{Top vs.\ Middle} targets critics sensitive to subtle distinctions. For each channel $\chi$ and sensitivity level $\tau$, we select the highest-accuracy critic on validation task $\mathcal{T}_{\chi,\tau}^{\text{val}}$:
\begin{align}
c^*_{\chi,\tau} = \operatorname*{argmax}_{c\in\mathcal{C}_{\text{pool}}} \operatorname{Acc}(c\mid\mathcal{T}_{\chi,\tau}^{\text{val}}),
\end{align}
This yields the specialized pool $\mathcal{C}_{\text{task}}=\bigcup_{\chi,\tau}\{c^*_{\chi,\tau}\}$. To evaluate whether this critic pool is well-aligned and diverse to represent a broader pool, we compare it against both the ensemble mean of all critics and the top-performing single-critic pool.

\begin{wraptable}{r}{0.51\columnwidth}
\vspace{-20pt}
\centering
\caption{Held-out test accuracy for writing critics.}
\label{tab:test-critics}
\scriptsize
\begin{tabular}{lccc}
\toprule
Comparison & Mean Critic & Single Best & Task-Wise Best \\
\midrule
Top vs.\ Middle & 0.557 & 0.554 & \textbf{0.578} \\
Top vs.\ Bottom & 0.654 & 0.670 & \textbf{0.716} \\
\bottomrule
\end{tabular}
\vspace{-10pt}
\end{wraptable}

Notably, task-specific selection improves held-out accuracy over the mean and single-best baselines (Table~\ref{tab:test-critics}). We provide channel-specific validation results in Table~\ref{tab:val-breakdown} of Appendix~\ref{sec:appendix_critic_selection}, which also follow a similar trend.

In Table~\ref{tab:val-in-context}, we demonstrate the impact of in-context sample size on the performance of selected writing critics. Specifically, we analyze how the number of tier-labeled samples provided to $p_{\text{script}}$ affects its ability to yield high-performing critics by evaluating settings with 0 (zero-shot), 15, and 45 samples across the selection strategies. Although even the zero-shot setting achieves favorable accuracy, providing 45 samples yields the best overall performance. Note that across all sample sizes, \textit{Task-Wise Best} remains the strongest approach.

\begin{table}[ht]
\centering
\vspace{-10pt}
\caption{Writing-critic validation accuracy as a function of in-context sample size. Task-wise selection has the highest reported accuracy for each sample size.}
\label{tab:val-in-context}
\small
\begin{tabular}{llccc}
\toprule
\# Samples & Comparison & Mean Critic & Single Best & Task-Wise Best \\
\midrule
\multirow{2}{*}{0}
& Top vs.\ Middle & 0.542 & 0.572 & 0.642 \\
& Top vs.\ Bottom & 0.700 & 0.728 & 0.802 \\
\midrule
\multirow{2}{*}{15}
& Top vs.\ Middle & 0.547 & 0.590 & 0.644 \\
& Top vs.\ Bottom & 0.697 & 0.702 & 0.808 \\
\midrule
\multirow{2}{*}{45}
& Top vs.\ Middle & 0.542 & 0.572 & 0.644 \\
& Top vs.\ Bottom & 0.708 & 0.740 & 0.830 \\
\bottomrule
\end{tabular}
\vspace{-10pt}
\end{table}

\subsection{Script Writing Loop}
\label{sec:writing}

\begin{figure}[t]
\centering
\includegraphics[width=\linewidth]{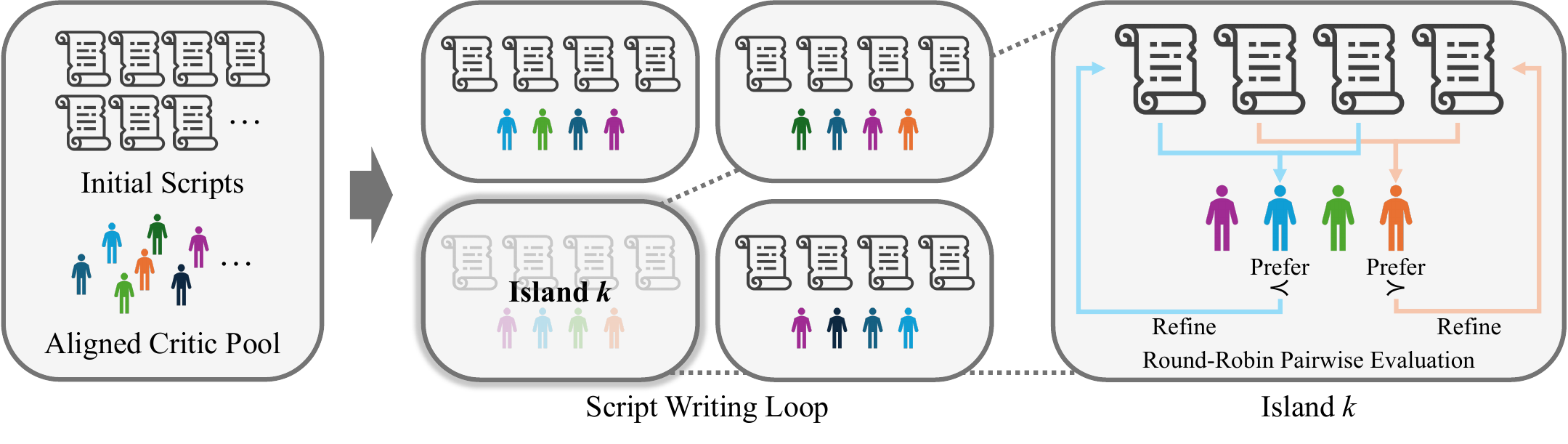}
\vspace{-25pt}
\caption{Script writing loop. Isolated script populations evolve on separate islands under distinct writing-critic committees sampled from the aligned pool. Losing scripts are refined through round-robin pairwise tournaments.}
\vspace{-10pt}
\label{fig:writing_archipelago}
\end{figure}

\paragraph{Islands and evolving fitness landscapes.}
We introduce an approach to iteratively evolve a population of scripts. Pairs of scripts are compared by a writing critic, which provides feedback to revise the weaker script. As the population evolves, weaker scripts are iteratively refined using the feedback, continuously raising the competitive baseline. A script that wins at generation $g$ may lose at $g+1$ not because it degraded, but because competitors improved. Defining fitness $f^{(g)}(s)$ as the expected win rate of script $s$ against the current population given a writing-critic committee $\mathcal{C}$ and script population at the current generation $\mathcal{S}^{(g)}$,
\begin{align}
f^{(g)}(s) = \mathbb{E}_{s'\sim\mathcal{S}^{(g)},c\sim\mathcal{C}} \bigl[\mathbb{I}\bigl[c(s,s')\mapsto(s,\cdot)\bigr]\bigr],
\end{align}
this formulation ideally implements a competitive environment that grows more demanding, \ie, $\mathbb{E}[f^{(g)}(s)] \geq \mathbb{E}[f^{(g+1)}(s)]$, forcing continuous adaptation.

To encourage diversity of solutions, we partition the global script population into $K$ isolated islands $\{I_1,\ldots,I_K\}$, each governed by a specialized writing-critic committee $\mathcal{C}_k$ drawn from $\mathcal{C}_\text{task}$ in Section~\ref{sec:critic}. These separate committees embody distinct comedic preferences, while being aligned with engagement patterns. The fitness landscape on island $k$ is shaped by two coupled elements: (1) the island-specific writing-critic committee $\mathcal{C}_k$, which defines evaluative standards, and (2) the evolving script population $\mathcal{S}_k$, which determines the comparative baseline. Because both critics and populations differ across islands, the fitness landscapes tend to diverge, yielding a Pareto frontier of diverse comedic styles. Figure~\ref{fig:writing_archipelago} illustrates this island structure and the critic-guided competition.

\paragraph{Pairwise evaluation with critic-guided update.}
Within each island, evolution proceeds through round-robin pairwise evaluation. At each iteration, two scripts $s_i, s_j\in\mathcal{S}_k$ are compared by every critic in the local committee. Each critic $c_e\in\mathcal{C}_k$ performs an independent evaluation:
\begin{align}
c_e(s_i,s_j) \;\mapsto\; (w_{c_e},\,\phi_{c_e}).
\end{align}
Let $s_\ell\in\{s_i,s_j\}\setminus\{w_{c_e}\}$ denote the losing script. It undergoes a \emph{critic-guided update} driven by the feedback:
\begin{align}
s_\ell \;\leftarrow\; U(s_\ell,\,\phi_{c_e}),
\end{align}
where $U\colon\mathcal{S}\times\Phi\to\mathcal{S}$ is the update operator that rewrites the script according to natural-language feedback $\phi_{c_e}$, $\mathcal{S}$ denotes the space of scripts, and $\Phi$ denotes the space of natural-language feedback. Note that we only update the losing script. This compact operator integrates two classical evolutionary mechanisms in a single call: the comparative feedback $\phi_{c_e}$ encourages the loser to incorporate the winner's strengths, resulting in semantic \emph{crossover} that transfers beneficial features from the superior script, while $U$ simultaneously introduces semantic \emph{mutation} by rewriting the script under critic guidance, exploring variations that may uncover novel comedic approaches.

\begin{figure}[t]
\centering
\includegraphics[width=\linewidth]{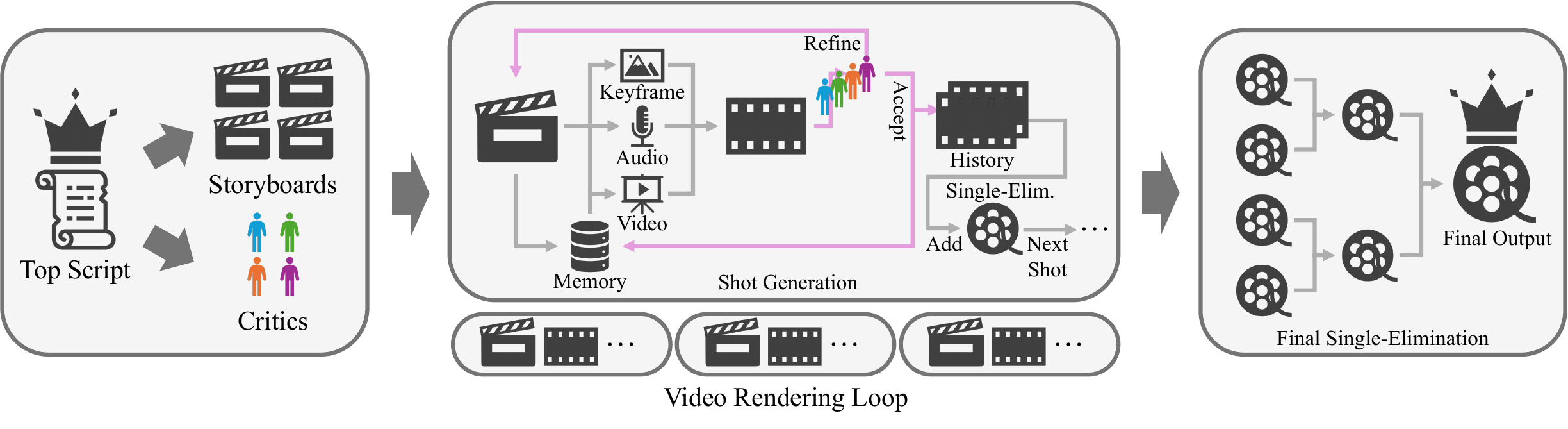}
\vspace{-25pt}
\caption{Video rendering loop. Script-conditioned rendering critics refine scene directions, while shot- and video-level tournaments select from refinement histories and scene realizations, \ie, depth and breadth, respectively.}
\vspace{-10pt}
\label{fig:rendering_archipelago}
\end{figure}

\subsection{Video Rendering Loop}
\label{sec:rendering}
The rendering stage translates selected scripts into videos using script-conditioned critics and competition-based refinement. Figure~\ref{fig:rendering_archipelago} shows a rough overview of this process.

\paragraph{Script-conditioned rendering-critic generation.}
Given a refined script $s$, a video meta-critic $p_{\text{render}}$ generates a script-conditioned set of rendering critics:
\begin{equation}
\mathcal{C}_{\text{render}}\sim p_{\text{render}}(\mathcal{C}\mid s).
\end{equation}
Each critic provides a different lens for visually realizing the narrative.

\paragraph{Storyboarding.}
Because video generation is computationally expensive, we first storyboard the visual content conditioned on the script. For each script $s$, a scene director generates $D$ scene directions $\{d^{(1)},\ldots,d^{(D)}\}$, where each $d^{(j)}=\{d^{(j)}_1,\ldots,d^{(j)}_{N_j}\}$ specifies $N_j$ shots in text form. Each shot specification defines reference characters, backgrounds, and previous shots, along with poses, expressions, camera framing, or actions. A structured memory bank $\mathcal{M}$ stores character assets, backgrounds, and the last frame of each finalized shot so subsequent specifications can preserve continuity. In addition, we use chain-of-thought reasoning by introducing setup notes to help the director plan complex visual arrangements before producing shot specifications.

\paragraph{Iterative shot refinement with history tournament.}
For direction $d^{(j)}$ and shot $i$, we render an initial shot and iteratively revise its specification:
\begin{align}
v_i^{(0)} \;\leftarrow\; \mathrm{Render}\bigl(d_i^{(j,0)},\,V_{<i},\,\mathcal{M}\bigr),
\end{align}
where $\mathrm{Render}$ includes image, voice, and video generation based on the direction. For $m=0,\ldots,|\mathcal{C}_{\text{render}}|-1$, a rendering critic evaluates the current shot and proposes a refined specification:
\begin{align}
c\bigl(d_i^{(j,m)},\,V_{<i},\,v_i^{(m)}\mid s\bigr) \;\mapsto\; \bigl(d_i^{(j,m+1)},\,\phi_{\text{refine}}\bigr),
\end{align}
where $V_{<i}=[v_1,\ldots,v_{i-1}]$ are previously finalized shots. The updated specification sequentially guides the next render.

Refinement accumulates a history $\mathcal{H}_i^{(j)} = \{v_i^{(0)},\ldots,v_i^{(|\mathcal{C}_{\text{render}}|)}\}$. Rather than accepting the final iteration, we tournament the refinement history $\mathcal{H}_i^{(j)} = \{v_i^{(0)},\ldots,v_i^{(|\mathcal{C}_{\text{render}}|)}\}$ and select the winner:
\begin{align}
v_i^* = \mathrm{SingleElimination}\!\bigl(\mathcal{H}_i^{(j)},\,\mathcal{C}_{\text{render}},\,s,\,V_{<i}\bigr).
\end{align}
This guards quality against degradation after refinement. The history tournament selects with respect to \emph{depth}, encouraging exploitation of the given scene direction.

\paragraph{Test-time scaling via scene-level tournament.}
After rendering the $D$ scene directions into complete videos $\{\mathcal{V}^{(1)},\ldots,\mathcal{V}^{(D)}\}$, where each $\mathcal{V}^{(j)}=[v^*_1,\ldots,v^*_{N_j}]$ consists of selected shots, we perform a final tournament:
\begin{align}
\mathcal{V}^* = \mathrm{SingleElimination}\!\bigl(\{\mathcal{V}^{(1)},\ldots,\mathcal{V}^{(D)}\},\,\mathcal{C}_{\text{render}},\,s\bigr).
\end{align}
Scaling $D$ at inference time directs additional compute toward rendering quality without retraining, providing a parallelizable test-time scaling axis. The scene-level tournament selects across \emph{breadth}, encouraging exploration of diverse scene realizations.

\begin{table}[t]
\centering
\caption{Human evaluation of generated videos. Scores are means on 7-point scales. \textit{vs.\ Human} compares each average, non-curated output with professional, human-made sketch comedy.}
\label{tab:method_comparison}
\resizebox{\columnwidth}{!}{%
\begin{tabular}{lccccccc}
\toprule
Method & \makebox[1.7cm]{Funniness↑} & \makebox[1.7cm]{Watch More↑} & \makebox[1.7cm]{vs. Human↑} & \makebox[1.7cm]{Script↑} & \makebox[1.7cm]{Narrative↑} & \makebox[1.7cm]{Realism↑} & \makebox[1.7cm]{Consistency↑} \\
\midrule
\textcolor{gray}{Veo 3.1}~\citep{google2025veo} & \textcolor{gray}{2.32} & \textcolor{gray}{2.36} & \textcolor{gray}{2.27} & \textcolor{gray}{2.18} & \textcolor{gray}{3.32} & \textcolor{gray}{4.91} & \textcolor{gray}{5.05} \\
\textcolor{gray}{Sora 2}~\citep{openai2025sora} & \textcolor{gray}{2.73} & \textcolor{gray}{2.73} & \textcolor{gray}{2.32} & \textcolor{gray}{2.45} & \textcolor{gray}{3.36} & \textcolor{gray}{5.73} & \textcolor{gray}{5.50} \\
VGoT~\citep{zheng2024videogen} & 1.18 & 1.27 & 1.14 & 1.00 & 1.23 & 2.00 & 2.32 \\
MovieAgent~\citep{wu2025automated} & 1.27 & 1.09 & 1.18 & 1.09 & 1.09 & 1.27 & 1.14 \\
COMIC (Ours) & \textbf{3.45} & \textbf{3.09} & \textbf{3.05} & \textbf{3.32} & \textbf{4.50} & \textbf{4.27} & \textbf{4.50} \\
\bottomrule
\end{tabular}
}
\end{table}

\begin{table}[t]
\centering
\vspace{-10pt}
\caption{Automated evaluation by benchmark video critics independently sampled and separately selected from COMIC's optimization critics, averaged across channels. \textit{Single Best} uses one top benchmark critic; \textit{Channel-Wise Best} aggregates per-channel benchmark critics.}
\label{tab:critic-metrics}
\resizebox{\columnwidth}{!}{%
\begin{tabular}{lcccccc}
\toprule
\multirow{2}{*}{Method} & \multicolumn{3}{c}{Single Best} & \multicolumn{3}{c}{Channel-Wise Best} \\
\cmidrule(lr){2-4} \cmidrule(lr){5-7}
 & Win Rate & Inter-Diversity & Intra-Diversity & Win Rate & Inter-Diversity & Intra-Diversity \\
\midrule
Veo 3.1~\citep{google2025veo}  & 0.010 & 0.308 & 0.369 & 0.105 & 0.263 & 0.360 \\
Sora 2~\citep{openai2025sora} & \underline{0.075} & \underline{0.531} & \textbf{0.722} & \underline{0.175} & \underline{0.310} & \underline{0.563} \\
VGoT~\citep{zheng2024videogen} & 0.000 & 0.000 & 0.000 & 0.010 & 0.105 & 0.189 \\
MovieAgent~\citep{wu2025automated} & 0.000 & 0.000 & 0.000 & 0.130 & 0.088 & 0.180 \\
COMIC (Ours) & \textbf{0.440} & \textbf{0.780} & \underline{0.682} & \textbf{0.390} & \textbf{0.519} & \textbf{0.693} \\
\bottomrule
\end{tabular}
}
\vspace{-10pt}
\end{table}

\begin{figure}[t]
\centering
\includegraphics[width=0.32\linewidth]{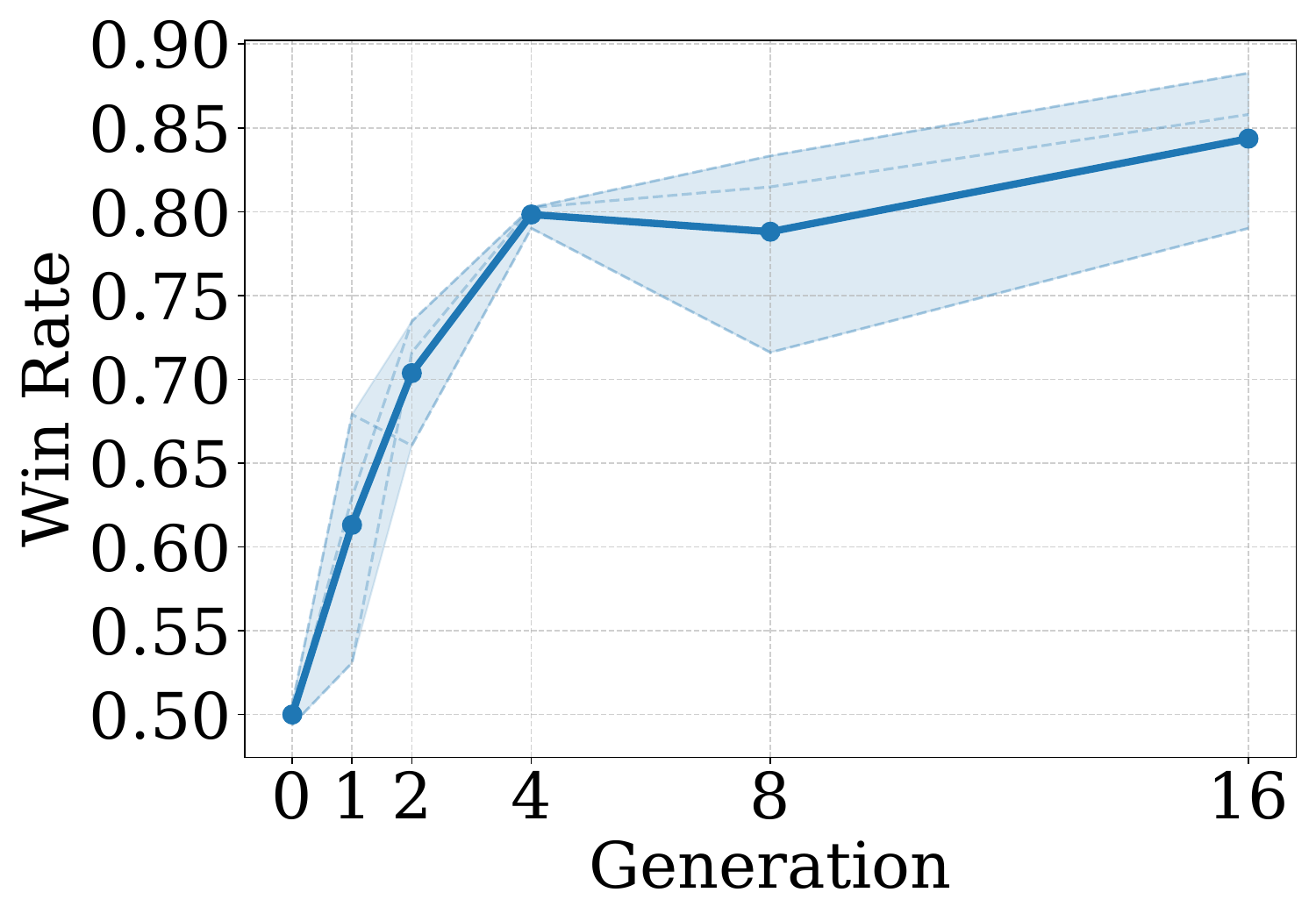}
\includegraphics[width=0.32\linewidth]{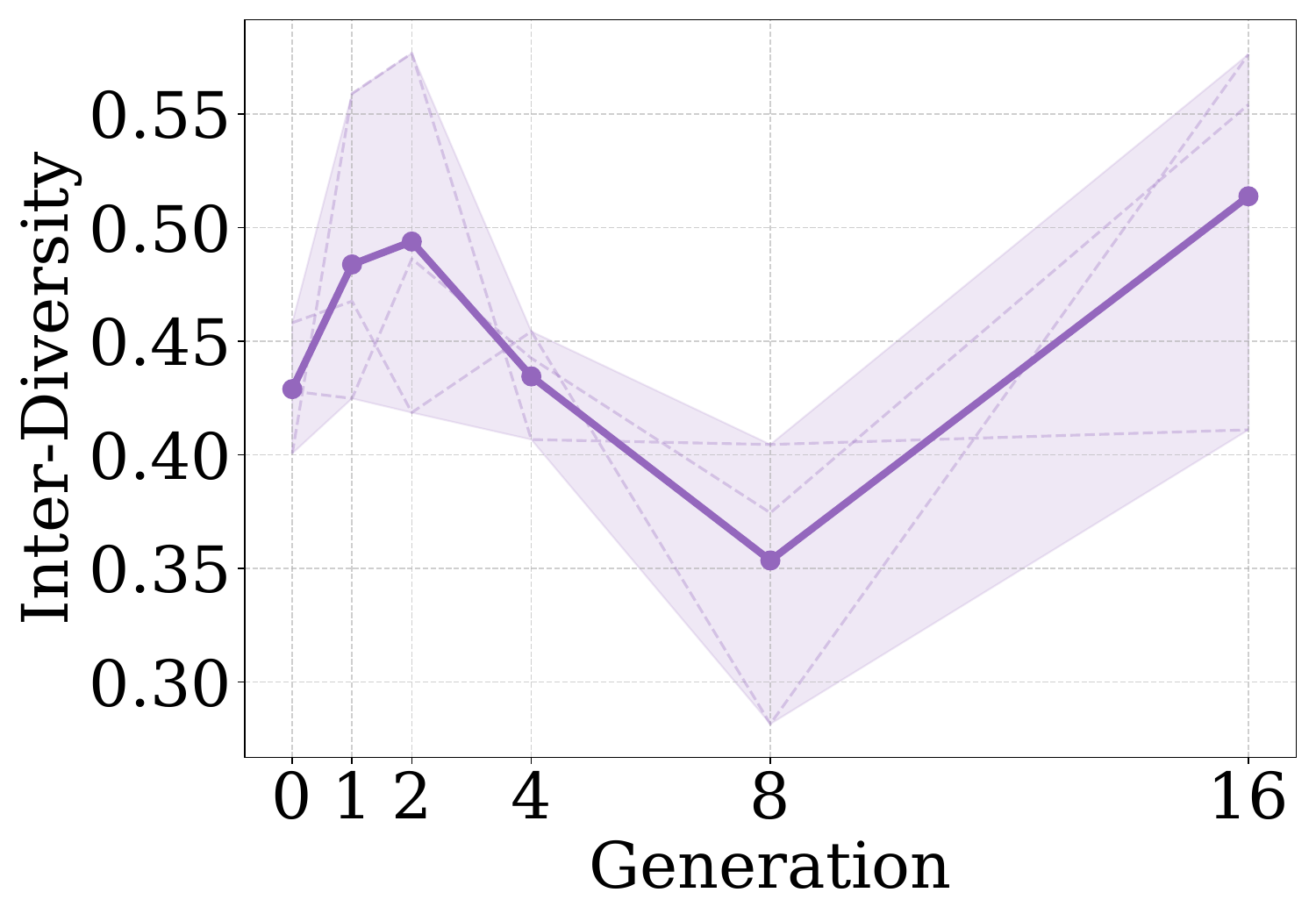}
\includegraphics[width=0.32\linewidth]{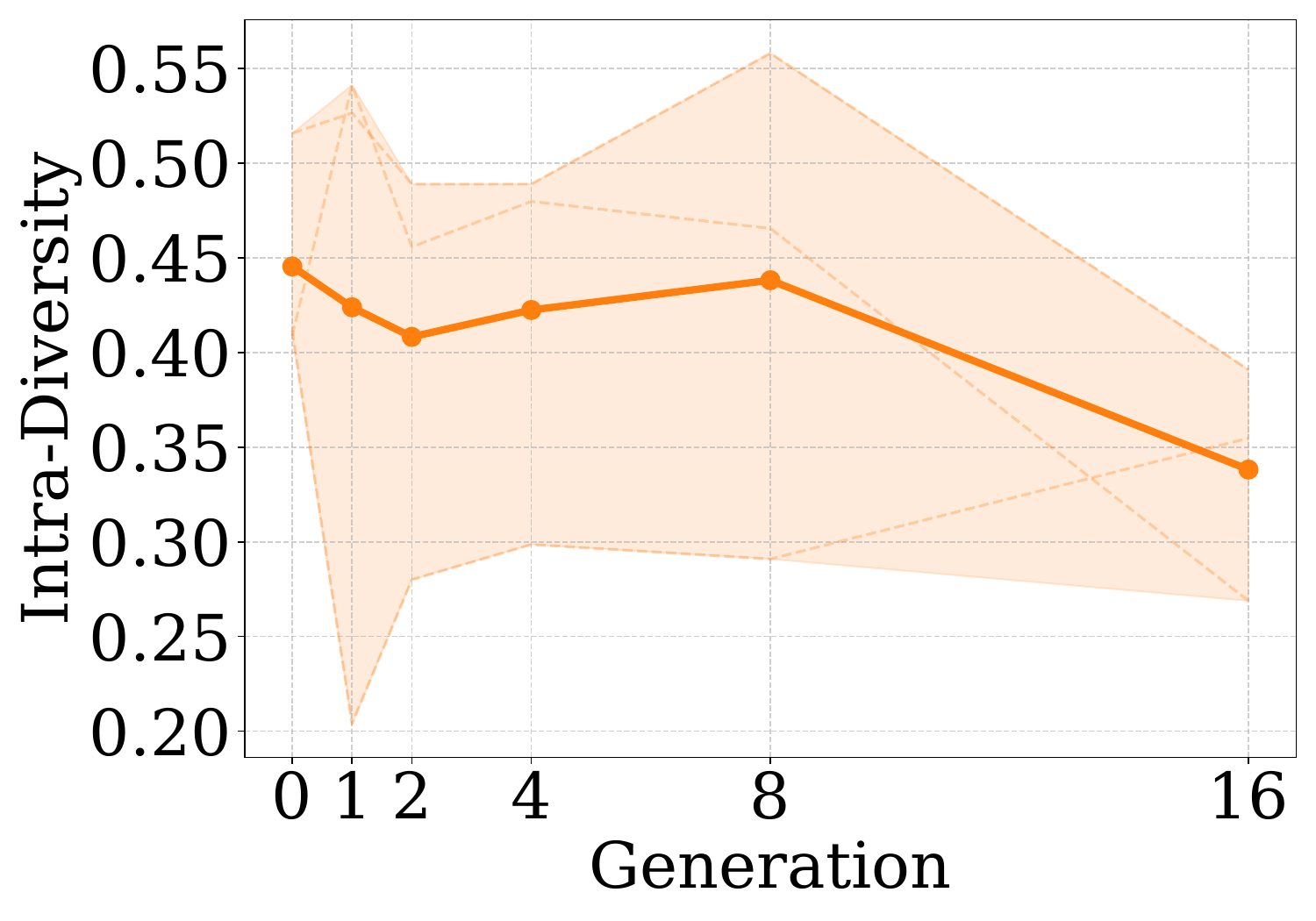}
\vspace{-15pt}
\caption{Effect of generation, measured by \textit{Win Rate}, \textit{Inter-Diversity}, and \textit{Intra-Diversity} across iterations relative to the 0th generation. Average values are presented as solid lines, and the range is depicted as shading. Through the 4th generation, the win rate increases substantially. \textit{Inter-Diversity} (diversity across scripts) initially drops due to the emergence of coherent, favorable responses but increases as generations progress, driven by our divergent mechanism.}
\vspace{-10pt}
\label{fig:self_improvement}
\end{figure}

\begin{figure}[t]
\centering
\includegraphics[width=\linewidth]{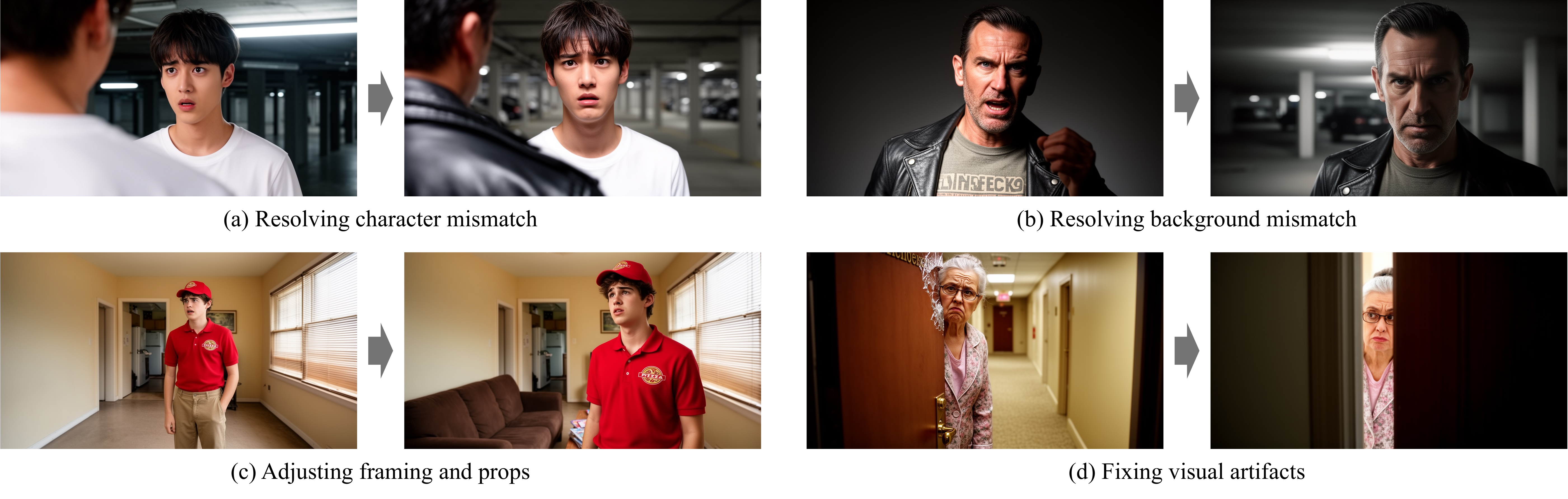}
\vspace{-25pt}
\caption{Qualitative examples of corrections made by script-conditioned rendering critics, including character, background, framing, and visual-artifact fixes.}
\vspace{-10pt}
\label{fig:video_critics}
\end{figure}

\section{Experiments}

\subsection{Implementation Details}
COMIC exposes several scaling dimensions: the number of islands $K$, scripts per island $|\mathcal{S}_k|$, writing critics per island $|\mathcal{C}_k|$, scene directions $D$, and rendering critics $|\mathcal{C}_{\text{render}}|$. We define small, base, and large configurations across these dimensions. Unless otherwise noted, we report results from the 4th generation and use the base configuration. The base configuration runs in approximately one day on a single GPU with an API budget of around \$5. Model details and full configurations are in Appendix~\ref{sec:appendix_models} (Tables~\ref{tab:models} and \ref{tab:scales}).

We evaluate scripts and videos by pairwise comparisons. Let $A$, $B$, and $E$ denote reference samples, generated samples, and evaluators, and let $P_{e,b,a}$ be the probability that $b$ beats $a$ under $e$. \textit{Win Rate} measures average quality, $Q_{\text{avg}}=\mathbb{E}_{e,b,a}[P_{e,b,a}]$, where values above 0.5 favor generated samples. \textit{Inter-Diversity} measures variation across generated samples, $D_{\text{inter}}=\mathbb{E}_{e,a}[\text{Var}_b(P_{e,b,a})]/(Q_{\text{avg}}(1-Q_{\text{avg}}))$. \textit{Intra-Diversity} measures evaluator/reference variation for each generated sample, $D_{\text{intra}}=\mathbb{E}_{b}[\text{Var}_{e,a}(P_{e,b,a})]/(Q_{\text{avg}}(1-Q_{\text{avg}}))$. The normalization makes both diversity metrics scale-invariant.

\subsection{Video Results}

The Figure~\ref{fig:teaser} teaser illustrates sketch comedy videos generated by COMIC from minimal specifications (\eg, a portrait, voice sample, and brief description). Starting from these inputs, the system autonomously develops comedic arcs with setups, escalating tension, and payoffs. The sketches span dry, deadpan, and surreal styles while maintaining character, background, and shot continuity; additional videos are provided in Figure~\ref{fig:more_examples} in Appendix~\ref{sec:appendix_video_results}, and the supplementary material.

\subsection{Baseline Comparison}
\label{sec:baseline_comparison}

We compare COMIC against VideoGen-of-Thought~\citep{zheng2024videogen} and MovieAgent~\citep{wu2025automated} as agentic video production baselines. These represent storyboard-driven long-form agentic video generation but lack iterative refinement or competitive selection. We further compare against black-box frontier models Sora 2~\citep{openai2025sora} and Veo 3.1~\citep{google2025veo}, to assess the contribution of our agentic pipeline over raw generative capability.

\paragraph{Automated evaluation.}
For benchmarking, we prompt a video meta-critic agent to synthesize a pool of candidate benchmark critics with diverse personas. Crucially, this pool is independently sampled and its critics are separately selected from the writing and script-conditioned rendering critics. The selected benchmark critics compare generated videos with held-out middle-tier test videos representing ``median'' sketch comedies.

We consider two aggregation approaches: \textit{Single Best}, which selects the highest-accuracy critic on validation, and \textit{Channel-Wise Best}, which selects critics independently per channel to capture diverse comedic traditions. Table~\ref{tab:critic-metrics} reports win rate, inter-diversity, and intra-diversity averaged across channels. COMIC has the highest win rate under both aggregations, and also has the highest inter- and intra-diversity; the automated ordering (COMIC $>$ Sora $>$ Veo $>$ MA $\approx$ VGoT) agrees with the human evaluation results presented in the following paragraph.

\paragraph{Human evaluation.}
We conducted a blind, randomized human evaluation of five methods using automatically selected, non-curated COMIC outputs. Table~\ref{tab:method_comparison} reports mean scores on 7-point scales. COMIC receives the highest mean scores among the agentic methods across \textit{Funniness}, \textit{Watch More}, \textit{Script}, \textit{Narrative}, \textit{Realism}, and \textit{Consistency}. Although Sora 2 and Veo 3.1 score higher on \textit{Realism} and \textit{Consistency}, COMIC scores higher on \textit{Funniness}, \textit{Watch More}, \textit{vs.~Human}, \textit{Script}, and \textit{Narrative} while producing longer videos. For further details and the evaluation protocol, see Appendix~\ref{sec:appendix_human_eval}.

\paragraph{Comparison against human-produced content.}
On the \textit{vs.\ Human} dimension, COMIC's average non-curated outputs score 3.05, while the best agentic baseline achieves only 1.14. The reference point of 4 is intentionally demanding, representing the level of typical professional comedy shows that audiences consume and serving as our North Star rather than a mediocre AI generation baseline. Showing meaningful progress, COMIC comes closer to this point than the frontier video models and agentic baselines.

\begin{wraptable}{r}{0.60\columnwidth}
\vspace{-20pt}
\centering
\caption{Controlled-backbone automated evaluation. Starred rows indicate COMIC backbones. \textit{No-Refine} is an ablation of our framework that removes multi-island iterative refinement.}
\label{tab:controlled-baselines}
\scriptsize
\resizebox{\linewidth}{!}{%
\begin{tabular}{llcccc}
\toprule
Benchmark critic & Metric & \makebox[1.25cm]{VGoT$^*$} & \makebox[1.25cm]{MovieAgent$^*$} & \makebox[1.25cm]{No-Refine$^*$} & \makebox[1.25cm]{\textbf{COMIC (Full)}$^*$} \\
\midrule
\multirow{2}{*}{Single Best}
& Win Rate$\uparrow$ & 0.03 & 0.02 & 0.30 & \textbf{0.44} \\
& Inter-Div.$\uparrow$ & 0.62 & 0.46 & 0.70 & \textbf{0.78} \\
\midrule
\multirow{2}{*}{Channel-Wise Best}
& Win Rate$\uparrow$ & 0.17 & 0.13 & 0.31 & \textbf{0.39} \\
& Inter-Div.$\uparrow$ & 0.37 & 0.33 & 0.50 & \textbf{0.52} \\
\bottomrule
\end{tabular}%
}
\vspace{-10pt}
\end{wraptable}

\paragraph{Controlled backbones and refinement.}
We rerun VGoT and MovieAgent with the language, image, and video backbones used by COMIC. As shown in Table~\ref{tab:controlled-baselines}, our full pipeline consistently outperforms the controlled baselines in win rate and inter-diversity across both sets of benchmark critics.

\subsection{Ablation Study}

\paragraph{Island-based evolution and iterative refinement.} Figure~\ref{fig:self_improvement} tracks win rates and diversity across successive generations. The win rate rises sharply through generation 4, confirming that pairwise tournaments drive rapid improvement. Inter-population diversity initially drops as populations converge toward generally effective strategies, then recovers as distinct critic committees push them toward unique niches. Examples of script refinement are shown in Figure~\ref{fig:script-refinement} in Appendix~\ref{sec:script_refinement}. Figure~\ref{fig:video_critics} illustrates how rendering critics correct issues such as character mismatches and framing errors.

\begin{wrapfigure}{r}{0.25\columnwidth}
\vspace{-15pt}
\centering
\includegraphics[width=1.0\linewidth]{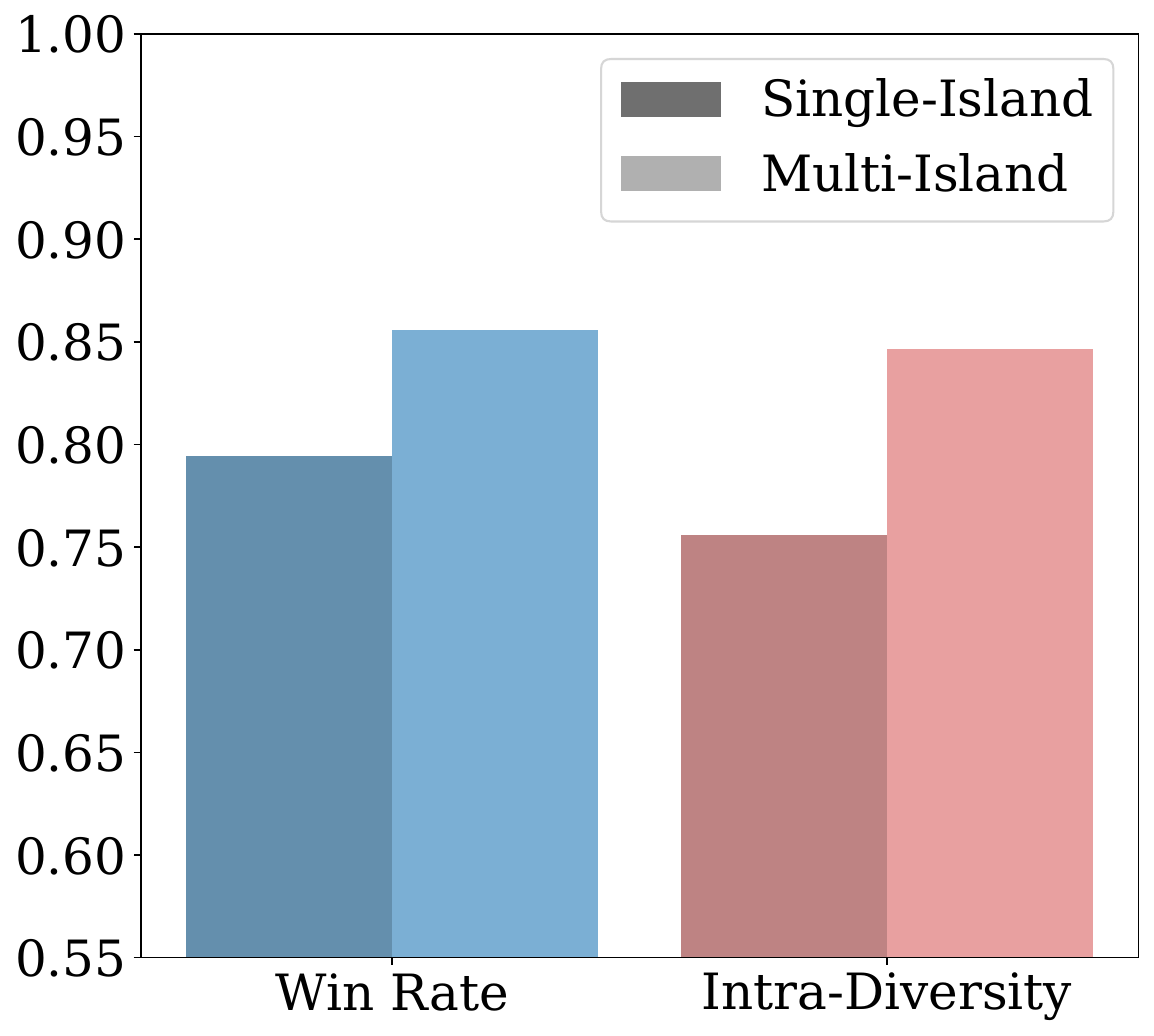}
\vspace{-25pt}
\caption{Single- and multi-island settings.}
\vspace{-20pt}
\label{fig:island_comparison}
\end{wrapfigure}

\paragraph{Single pool vs. multi-island.} We compare a single unified script pool with the multi-island topology while keeping the number of refinement iterations per script fixed. In the single-pool setting, every script is optimized under one shared population and selection pressure. In contrast, islands maintain separate populations and distinct writing-critic committees during local tournaments, allowing different comedic directions to develop before the final selection. As shown in Figure~\ref{fig:island_comparison}, the multi-island configuration improves both win rate and intra-diversity. Thus, preserving multiple local selection landscapes retains meaningfully distinct script trajectories without compromising average quality.

\begin{wrapfigure}{l}{0.25\columnwidth}
\vspace{-15pt}
\centering
\includegraphics[width=1.0\linewidth]{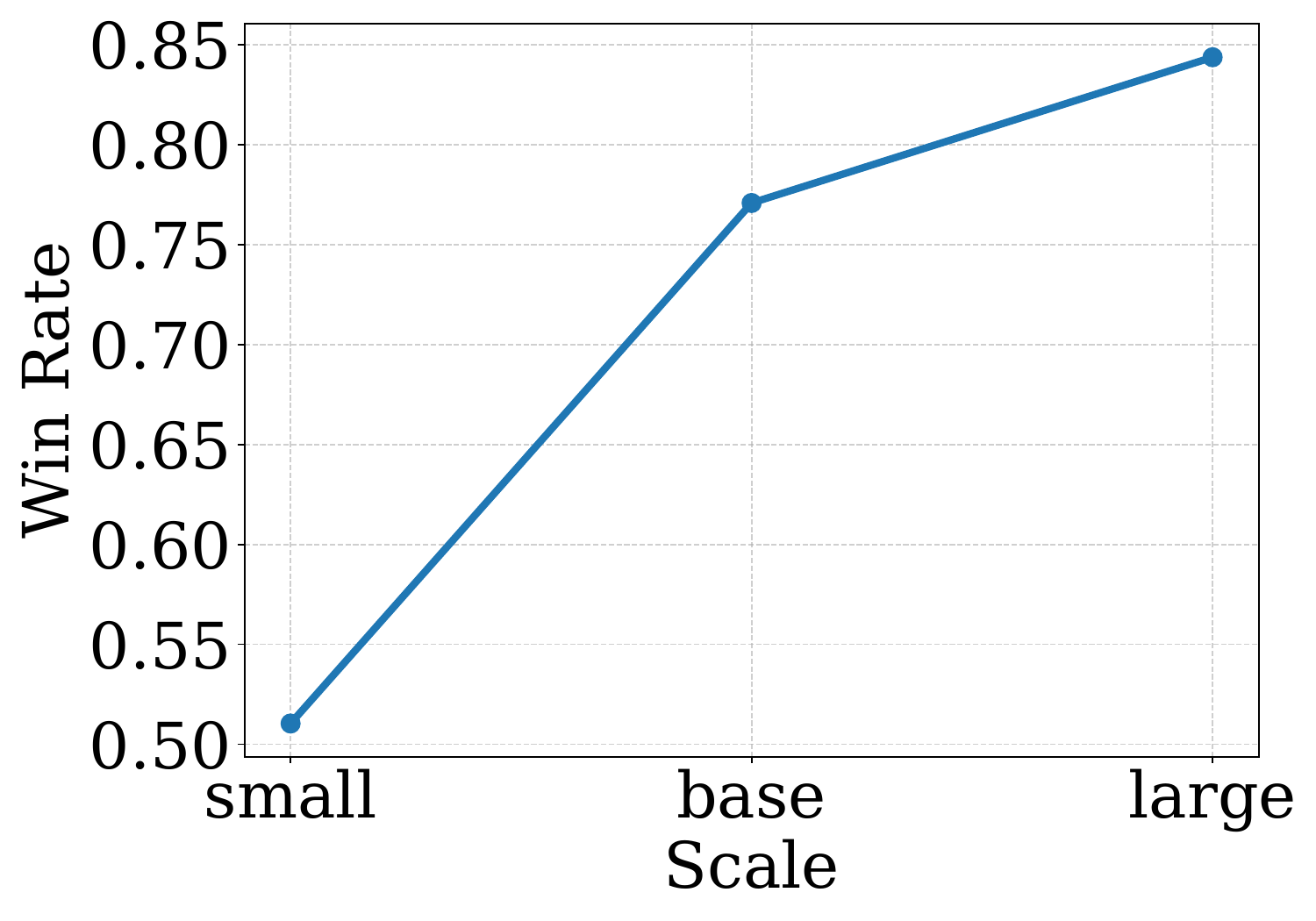}
\vspace{-25pt}
\caption{Scale.}
\vspace{-15pt}
\label{fig:scale}
\end{wrapfigure}

\paragraph{Scale.}
We compare small, base, and large configurations by jointly scaling the number of islands, scripts, and critics. Fig.~\ref{fig:scale} shows the win rate of each relative to the small scale. Performance improves consistently as the system is scaled. Top scripts from the large configuration achieve a higher win rate compared to the small and base baselines, showing that COMIC scales by trading test-time compute for enhanced performance. The full configurations are listed in Appendix~\ref{sec:scale_configurations}, Table~\ref{tab:scales}.

\begin{wrapfigure}{r}{0.45\columnwidth}
\vspace{-20pt}
\centering
\includegraphics[width=1.0\linewidth]{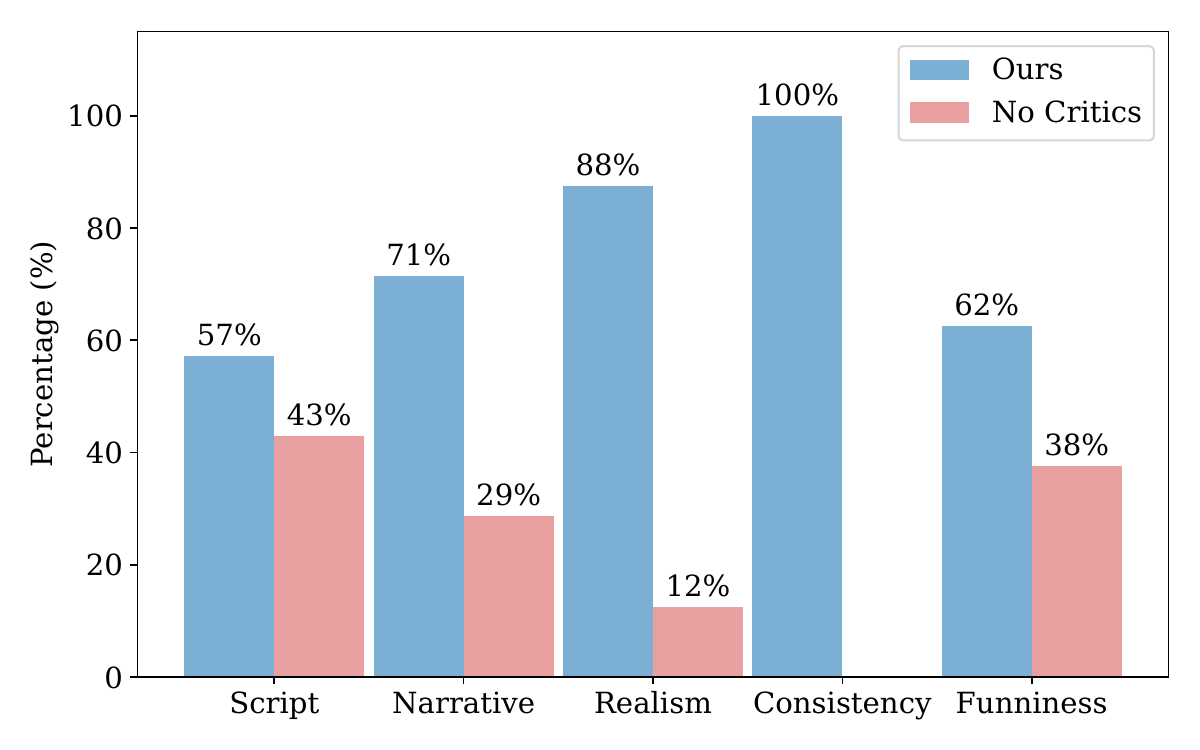}
\vspace{-25pt}
\caption{Human preference study against the critic-free ablation.}
\vspace{-10pt}
\label{fig:user_ablation}
\end{wrapfigure}

\paragraph{No-critics baseline.} Fig.~\ref{fig:user_ablation} reports an A/B preference study comparing the full COMIC pipeline with a critic-free ablation. Raters preferred the full system on \textit{Script}, \textit{Narrative}, \textit{Realism}, \textit{Consistency}, and \textit{Funniness}. The largest separations occur on visual consistency and realism, which are directly targeted by the script-conditioned rendering-critic loop, and the writing-oriented metrics also favor the full system. Additionally, automated results in Table~\ref{tab:controlled-baselines} show that COMIC also outperforms \textit{No-Refine} in both win rate and inter-diversity.

\section{Conclusion}
In this paper, we introduce COMIC, a fully automated multi-agent framework designed to tackle the extremely open-ended challenge of sketch comedy video generation. By shifting away from single-pass, fixed-objective pipelines, COMIC leverages a multi-island topology where diverse, human-aligned critic committees drive iterative refinement. The competitive pressure operates across both the narrative scriptwriting and visual rendering stages, allowing the system to explore a vast creative space while maintaining coherence. Our experiments demonstrate that COMIC outperforms existing agentic video baselines while offering a dual mechanism for test-time scaling. Ultimately, this work establishes a new state of the art for automated, engaging, long-form video production.

The iterative refinement process incurs computational costs, and normalized YouTube view counts can be noisy proxies for humor quality. Future work could add sound effects and quantify originality and attribution when building on large internet corpora.

COMIC's improvements emerge without parameter updates, gradient-based optimization, or a fixed reward signal, connecting to the \textit{Red Queen} hypothesis~\citep{van1973new} in evolutionary biology, wherein species must continuously evolve to maintain their fitness against co-evolving competitors. Unlike structured domains such as mathematics~\citep{woodruff2026accelerating} or board games~\citep{silver2016mastering}, comedy's shifting, context-dependent criteria make it a compelling proxy for open-ended, real-world problems. We believe that this work opens several directions for future research into other creative domains.

\bibliographystyle{iclr2027_conference}
\bibliography{main}

\newpage
\appendix

\begin{center}
{\Large \textbf{Appendix}}\vspace{10pt}
\end{center}

\section{Video Results}
\label{sec:appendix_video_results}

We include the MP4 video files as accompanying files and encourage readers to watch them.

\begin{figure}[htbp]
\centering
\includegraphics[width=0.95\linewidth]{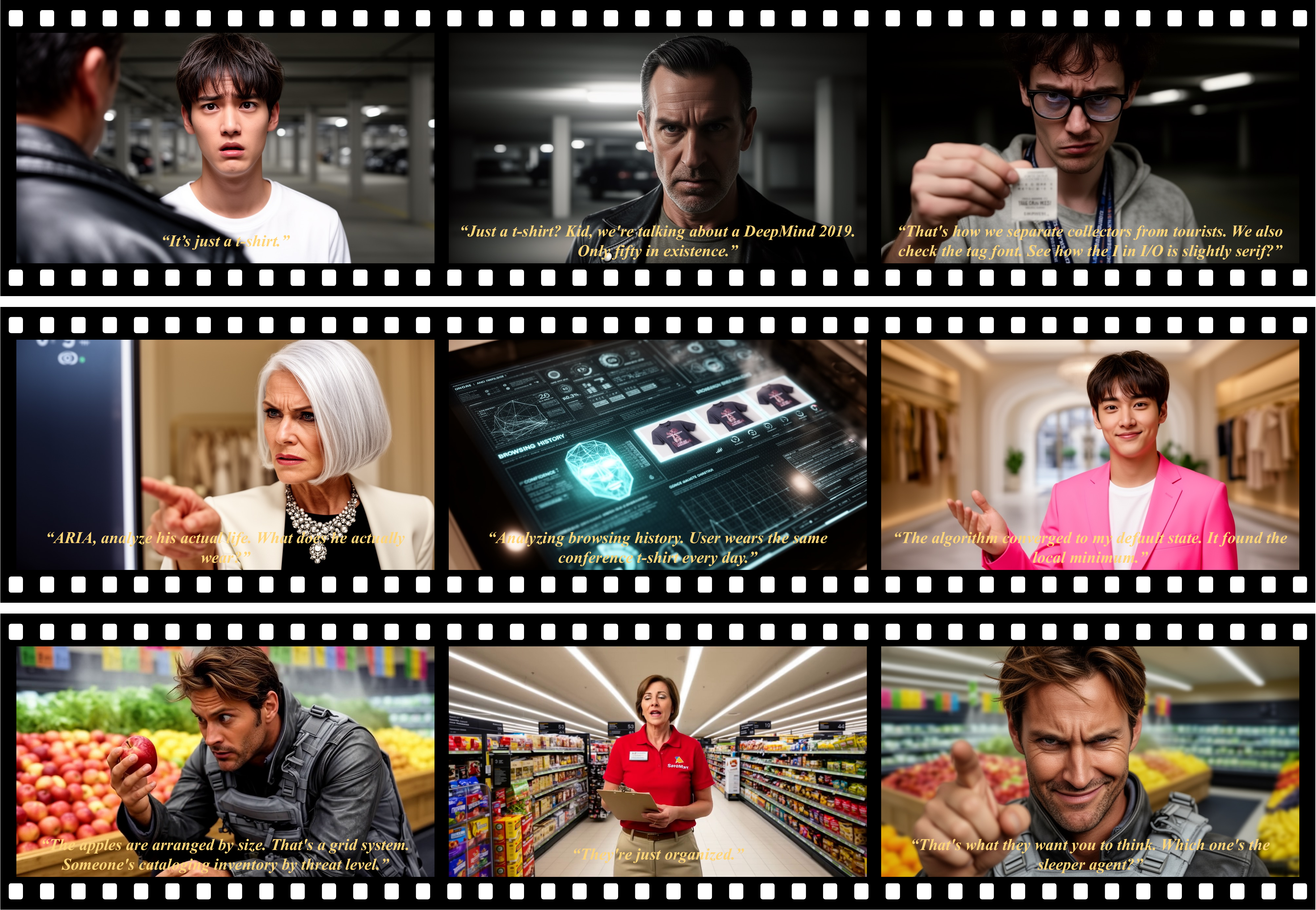}
\vspace{-10pt}
\caption{Additional sketch comedy videos featuring varied generated situations.}
\vspace{-10pt}
\label{fig:more_examples}
\end{figure}

\section{Writing-Critic Selection}
\label{sec:appendix_critic_selection}

\paragraph{Channel-specific validation.}
Table~\ref{tab:val-breakdown} reports writing-critic validation accuracy by channel and engagement tier, complementing the in-context sample-size comparison in the main paper (Table~\ref{tab:val-in-context}). Task-wise selection achieves the highest average accuracy for both comparisons.

\begin{table}[htbp]
\centering
\caption{Channel-specific writing-critic validation accuracy. Task-wise selection has the highest average accuracy across the reported channels and engagement tiers. FAH = Foil Arms \& Hog; VLDL = Viva La Dirt League; K\&P = Key \& Peele.}
\label{tab:val-breakdown}
\resizebox{\linewidth}{!}{%
\begin{tabular}{llcccccc}
\toprule
& Comparison & \makebox[1.2cm]{Studio C} & \makebox[1.2cm]{FAH} & \makebox[1.2cm]{VLDL} & \makebox[1.2cm]{K\&P} & \makebox[1.2cm]{SNL} & \textbf{Average} \\
\midrule
\multirow{2}{*}{Mean Critic}
& Top vs.\ Middle & 0.63 & 0.40 & 0.58 & 0.53 & 0.59 & 0.55 \\
& Top vs.\ Bottom & 0.72 & 0.62 & 0.79 & 0.60 & 0.81 & 0.71 \\
\midrule
\multirow{2}{*}{Single Best}
& Top vs.\ Middle & 0.62 & \textbf{0.52} & 0.56 & 0.57 & 0.59 & 0.57 \\
& Top vs.\ Bottom & 0.67 & 0.70 & 0.84 & 0.69 & 0.80 & 0.74 \\
\midrule
\multirow{2}{*}{Task-Wise Best}
& Top vs.\ Middle & \textbf{0.70} & \textbf{0.52} & \textbf{0.70} & \textbf{0.65} & \textbf{0.65} & \textbf{0.64} \\
& Top vs.\ Bottom & \textbf{0.85} & \textbf{0.78} & \textbf{0.85} & \textbf{0.72} & \textbf{0.95} & \textbf{0.83} \\
\bottomrule
\end{tabular}
}
\end{table}

\begin{figure}[htbp]
\centering
\includegraphics[width=0.9\linewidth]{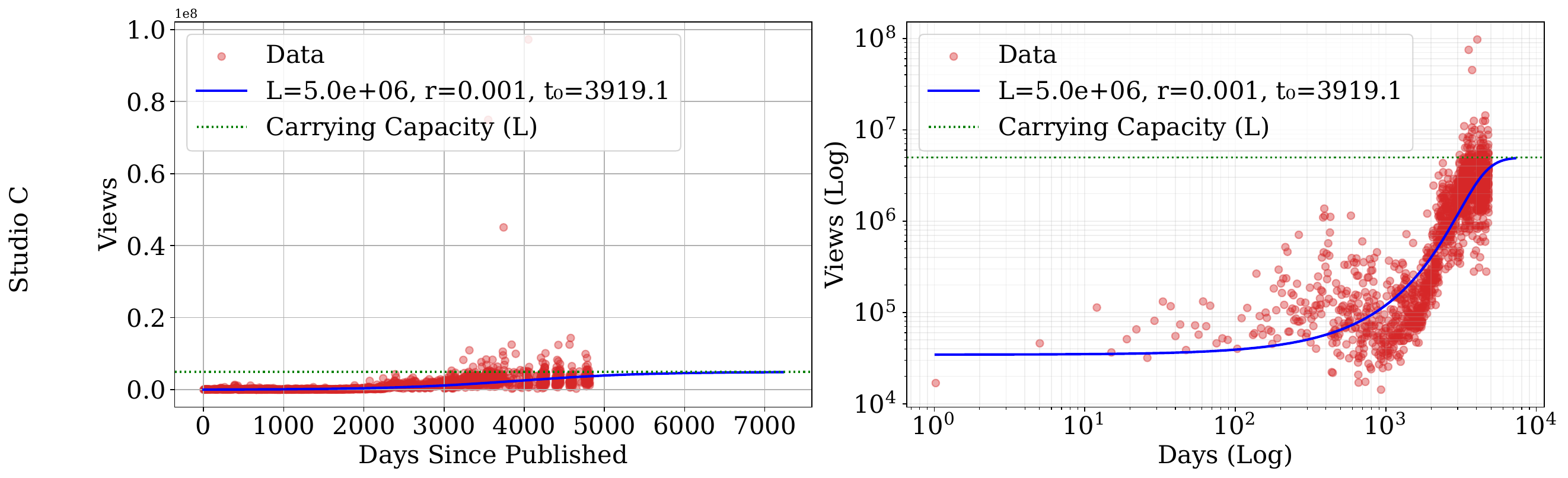}
\includegraphics[width=0.9\linewidth]{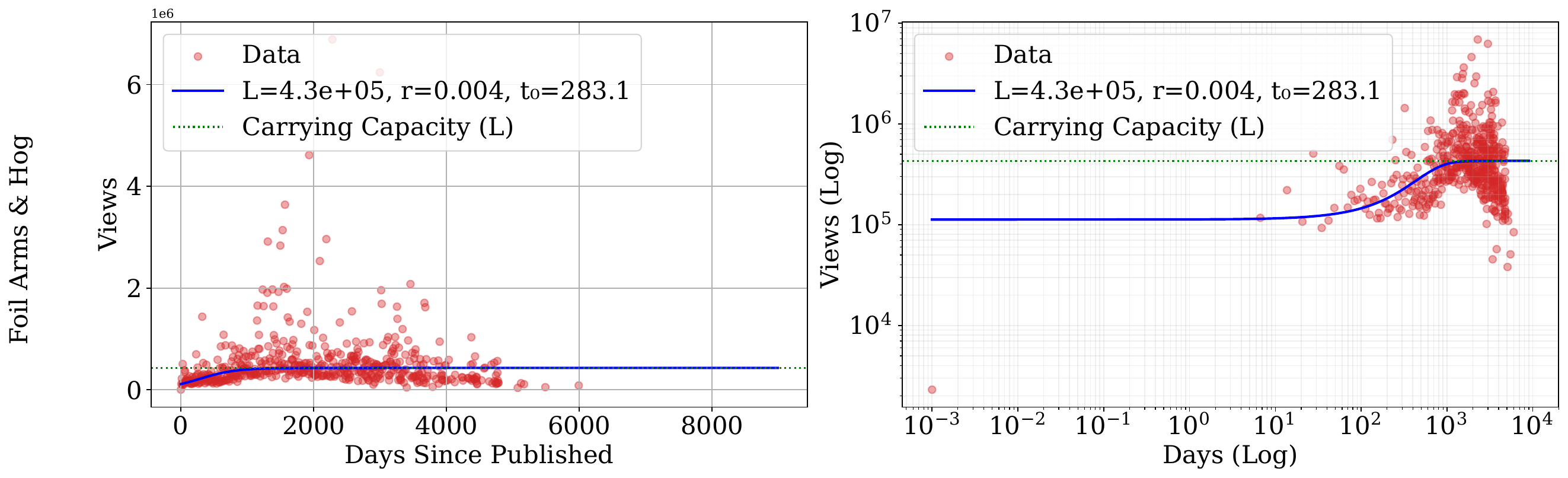}
\includegraphics[width=0.9\linewidth]{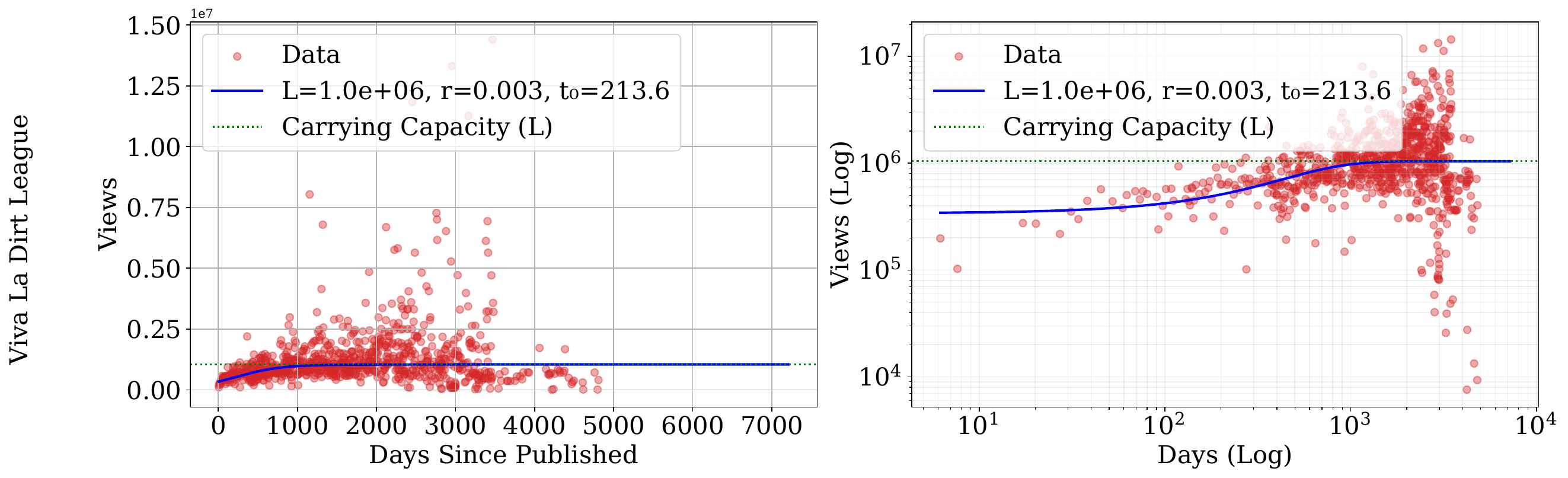}
\includegraphics[width=0.9\linewidth]{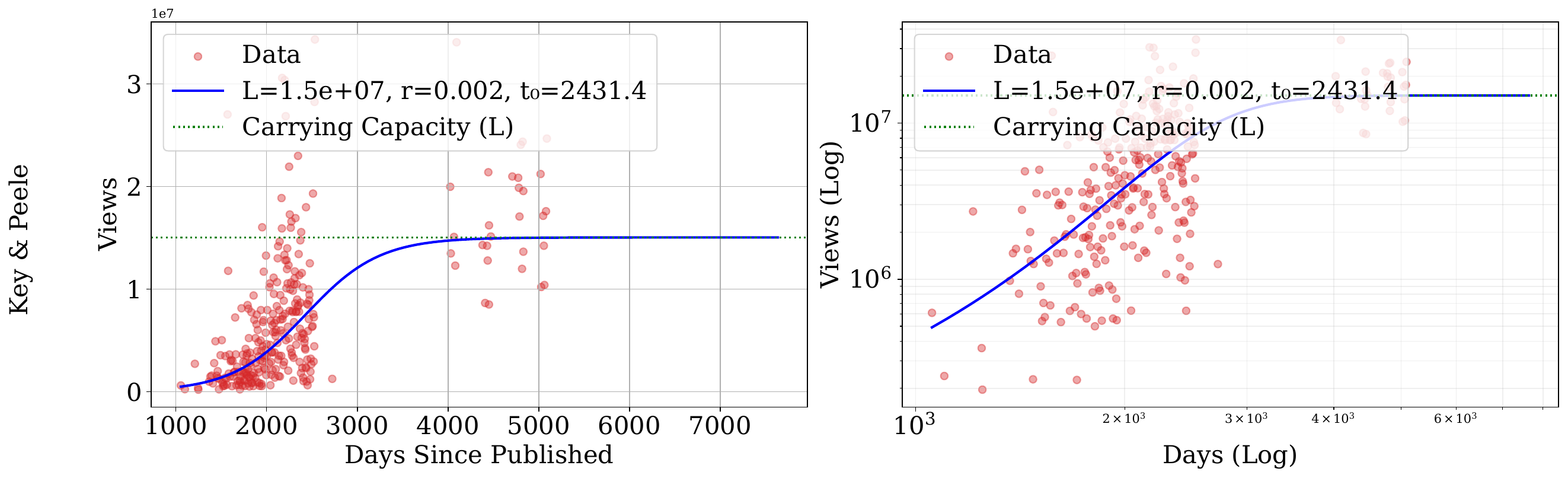}
\includegraphics[width=0.9\linewidth]{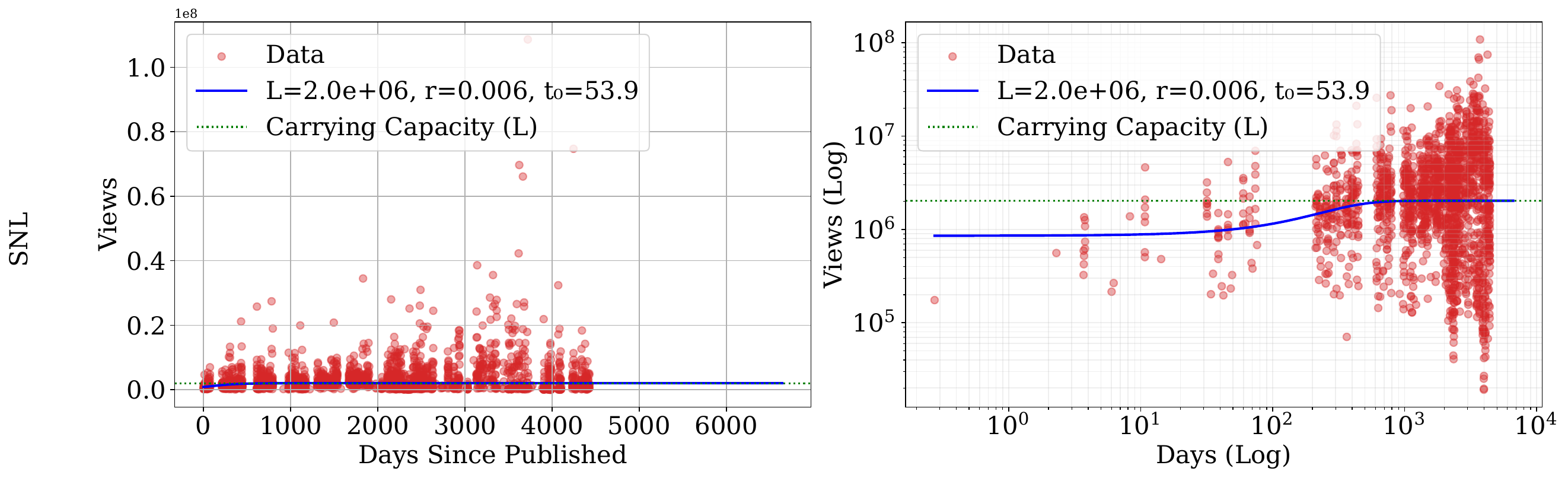}
\vspace{-10pt}
\caption{Logistic growth model fits for cumulative view counts across different comedy channels.}
\vspace{-10pt}
\label{fig:logistic}
\end{figure}

\paragraph{Data processing details.}
We collect 4,940 data points from five YouTube sketch comedy channels, excluding videos that do not meet our criteria for length and format. To construct channel-specific engagement scores, we model the cumulative view trajectory of each sketch using a logistic growth function. This reflects the S-shaped trajectories observed in online video view counts. Figure~\ref{fig:logistic} presents the fits of the logistic growth model for all five channels. Based on the projected infinite-horizon view count $L_{\text{proj}}$, we extract 30 data points from each of the five channels and each of the three tiers, resulting in 450 total data points. We then split them into $\mathcal{S}_{\text{in-context}}$, $\mathcal{S}_{\text{val}}$, and $\mathcal{S}_{\text{test}}$ as mentioned in the main paper.

\section{Models}
\label{sec:appendix_models}

A key feature of our framework is the ability to easily integrate modular foundation models at different production stages. Table~\ref{tab:models} summarizes the models used throughout the COMIC pipeline. We use Claude 4 Sonnet for concept and script generation. For tasks requiring efficiency during iterative refinement, we use Claude 4.5 Haiku for language-only writing critics and Gemini 3 Flash Preview for multimodal rendering and benchmark video critics. For tasks requiring more robust reasoning---such as script inspection, meta-critic instruction generation, and final scene direction---we employ Claude 4.5 Opus. We also use this model for the controlled backbone ablation.

For visual synthesis, we use FLUX.2 [dev]~\citep{flux-2-2025} to generate canonical character appearances and enhance visual clarity. For voice consistency, we employ ElevenLabs and Chatterbox-TTS~\citep{chatterboxtts2025} to generate stable voice prints for each character. These assets are stored in the visual memory bank $\mathcal{M}_{\text{visual}}$ and retrieved during shot generation. Video rendering leverages Wan 2.2~\citep{wan2025wan}. Since our image and video generation requires taking various modalities of input, \eg, image, text, and audio, we add a manifold-tangential guidance term to the predicted denoising score~\citep{cho2025tag}, which is agnostic to the input condition yet improves visual clarity and reduces hallucination. Additionally, we use an image critic to perform best-of-batch selection to further enhance quality and consistency. All image, video, and voice generation is performed on a single H200.

\begin{table}[htbp]
\centering
\caption{Models used in the pipeline.}
\label{tab:models}
\begin{tabular}{ll}
\toprule
\textbf{Component} & \textbf{Model} \\
\midrule
Concept and Script Writing & Claude 4 Sonnet \\
Language-Only Writing Critics & Claude 4.5 Haiku \\
Rendering / Benchmark Video Critics & Gemini 3 Flash Preview \\
Script Inspection & Claude 4.5 Opus \\
Scene Direction Generation & Claude 4.5 Opus \\
Meta-Critic & Claude 4.5 Opus \\
\midrule
Character Image Synthesis & FLUX.2 [dev] + TAG \\
Voice Synthesis & ElevenLabs \\
Voice Cloning & Chatterbox-TTS \\
Video Rendering & Wan 2.2 + TAG \\
\bottomrule
\end{tabular}
\end{table}

\begin{figure}[t]
\centering
\setlength{\fboxsep}{4pt}
\fbox{%
\begin{minipage}{0.96\linewidth}
\scriptsize\ttfamily
\setlength{\parskip}{2pt}
\begin{minipage}[t]{0.48\linewidth}
\textbf{Script 1 snippet}\par

Melissa Swift: Darling, this collection is revolutionary! The raw authenticity!

Haedol Park: Thanks, I guess? I just grab whatever's free at conferences.

Marcus Webb: Are we seriously doing a fashion shoot with a database shirt?

\scriptremove{Melissa Swift: Not databases, Marcus. Data liberation! Now drape it like liquid mercury.}

\scriptadd{Melissa Swift: Not databases, Marcus. Data liberation! This one's our cover piece.}

\scriptadd{Haedol Park: It's got a stain on it.}

\scriptadd{Melissa Swift: Vintage patina! We'll charge extra for authenticity!}

\scriptadd{Marcus Webb: That's marinara sauce.}

\scriptadd{Melissa Swift: Even better. Organic, handcrafted details. Now drape it like liquid mercury.}

\medskip\hrule\medskip

\textbf{Script 2 snippet}\par

Gloria: Your file is not red. No one has a red file.

\scriptremove{Eric Cross: One blink. Confirmed red. I'm going to need a new identity and a vehicle registered under an alias.}

\scriptadd{Eric Cross: One blink. Confirmed red. I need to see the DMV director.}

\scriptremove{Gloria: We can't do that.}

\scriptadd{Gloria: There is no director.}

\scriptremove{Eric Cross: Of course you can't. You're not authorized. Who is? Who's your supervisor?}

\scriptadd{Eric Cross: Exactly. That's what the director would say. What's behind that door?}
\end{minipage}%
\hfill
\begin{minipage}[t]{0.48\linewidth}
\textbf{Script 3 snippet}\par

\scriptremove{Campus Security: They're free. I get them at every event. I have eighty in my office. I use them for car wash rags.}

\scriptadd{Campus Security: They're free. I get them at every event. I have eighty in my office.}

\scriptadd{The Shirt Boss: You attend conferences?}

\scriptadd{Campus Security: No, I work the registration tables. They give them to me in bulk. I use them for car wash rags.}

The Shirt Boss: [staring] Car wash rags?

\scriptremove{Campus Security: The 2019 DeepMind one's particularly absorbent. You want some? [holds out a pile] I've got a whole filing cabinet.}

\scriptadd{Campus Security: The 2019 DeepMind one's particularly absorbent. You want some? [holds out a pile] I've got a whole filing cabinet. There's also a stack in the lost and found from 2017. Nobody claimed them.}

\scriptremove{[The Shirt Boss and Nervous Ned exit silently]}

\scriptadd{[The Shirt Boss and Nervous Ned exit silently, defeated]}

Haedol Park: What about the DeepMind?

\scriptremove{Campus Security: That one? Yeah, take it. Already used it twice.}

\scriptadd{Campus Security: That one? Yeah, take it. Already used it twice. [hands it over casually] Fair warning though---there's a coffee stain on the back.}
\end{minipage}
\end{minipage}%
}
\vspace{-5pt}
\caption{Script refinement snippets. Script 1, Script 2, and Script 3 show representative before/after edits. Red text is removed and green text is added by the iterative writing loop. Across snippets, refinement removes unnecessary dialogue, deepens the premise, and sharpens the payoff.}
\vspace{-10pt}
\label{fig:script-refinement}
\end{figure}

\section{Scale Configurations}
\label{sec:scale_configurations}

COMIC exposes several useful scaling dimensions. Round-robin pairwise evaluation within islands requires $\mathcal{O}(K|\mathcal{S}_k|^2|\mathcal{C}_k|)$ writing-critic calls per generation, which is substantially lower than a global tournament over all scripts that would require $\mathcal{O}(K^2|\mathcal{S}_k|^2|\mathcal{C}_k|)$ calls. Island-local evaluation is, therefore, not merely a diversity mechanism but also a computational necessity that makes iterative refinement at scale tractable. Table~\ref{tab:scales} summarizes the three scale configurations of COMIC.

\begin{table}[htbp]
\centering
\caption{Scale configurations for the COMIC framework.}
\label{tab:scales}
\begin{tabular}{lcccccc}
\toprule
Scale & $K$ & $|S_k|$ & $|\mathcal{C}_k|$ & $D$ & $|\mathcal{C}_\text{render}|$ \\
\midrule
Small & 2 & 2 & 2 & 1 & 0 \\
Base  & 3 & 3 & 3 & 2 & 1 \\
Large & 4 & 4 & 4 & 4 & 2 \\
\bottomrule
\end{tabular}
\end{table}

\paragraph{Computational complexity.}
We analyze the computational cost of COMIC across its key scaling dimensions: $K$ islands, $|S_k|$ scripts per island, $|\mathcal{C}_k|$ writing critics per island, $D$ scene directions, $|\mathcal{C}_\text{render}|$ rendering critics, maximum $N := \underset{j}{\max}\ N_j$ shots per video, and $G$ generations.

Within each island, a single generation requires a round-robin pairwise tournament over $|S_k|$ scripts, each evaluated by all $|\mathcal{C}_k|$ writing critics, incurring $\mathcal{O}(|S_k|^2 |\mathcal{C}_k|)$ evaluations per island. Across $K$ islands and $G$ generations, the total writing-stage cost is:
\begin{equation}
\mathcal{O}\!\left(G \cdot K \cdot |S_k|^2 \cdot |\mathcal{C}_k|\right).
\end{equation}
Crucially, this remains lower than a globally pooled tournament over all $K|S_k|$ scripts, which would require $\mathcal{O}(G \cdot K^2|S_k|^2 \cdot |\mathcal{C}_k|)$ evaluations.

For each selected script, we generate $D$ scene directions and refine each of the $N$ shots over $|\mathcal{C}_\text{render}|$ critic iterations, yielding $\mathcal{O}(D \cdot N \cdot |\mathcal{C}_\text{render}|)$ render calls per script. Shot-level tournament selection adds $\mathcal{O}(D \cdot N \cdot |\mathcal{C}_\text{render}|^2)$ comparisons (single-elimination), and the final video-level tournament over $D$ complete videos costs $\mathcal{O}(D \cdot |\mathcal{C}_\text{render}|)$ comparisons. The dominant term for the rendering stage is therefore:
\begin{equation}
\mathcal{O}\!\left(D \cdot N \cdot |\mathcal{C}_\text{render}|^2\right) \quad \text{render calls per script.}
\end{equation}

In practice, the base configuration takes approximately one day to run on a single H200 GPU with an API budget of around \$5, which is evidently lower than the cost of producing traditional comedy shows. Furthermore, the system is highly parallelizable. For example, we can parallelize API or model calls with respect to multiple islands and storyboards.

\section{Script Refinement}
\label{sec:script_refinement}
Figure~\ref{fig:script-refinement} visualizes representative changes made during iterative script refinement. The writing critics remove redundant exchanges, introduce new details, and strengthen character reactions to support comedic escalation.

\section{Storyboard Output Structure}
\label{sec:storyboard_structure}

The scene director agent outputs a structured JSON object that serves as the complete production specification for the video rendering stage. Given a script and user-specified character and background assets, the agent extracts all
characters and backgrounds, generates viewpoint variations for each base background (\eg, turning back, left, and right), and divides the script into shots. The full schema is shown in Listing~\ref{lst:storyboard}.

\begin{figure}[htbp]
\scriptsize
\begin{verbatim}
{
  "characters": [{
    "name": "...",            // Full name from script
    "description": "...",     // Physical and personality details
    "portrait_path": "...",   // Path to reference image, or null
    "voice_path": "...",      // Path to voice sample, or null
    "t2i_prompt": "...",      // Image gen prompt (if no portrait_path)
    "t2a_prompt": "...",      // Voice gen prompt (if no voice_path)
    "is_user_specified": true
  }],
  "backgrounds": [{
    "name": "...",            // Unique name, e.g., "Lab Front View"
    "description": "...",
    "image_path": "...",      // Path to reference image, or null
    "t2i_prompt": "...",      // Full scene prompt for base backgrounds
    "base_background": "...", // Parent background name (variations only)
    "variation_prompt": "...",// Viewpoint edit prompt (variations only)
    "is_user_specified": false
  }],
  "shots": [{
    "shot_id": "shot_01",
    "speaker": "...",         // Exact character name, or null
    "line": "...",            // Dialogue text
    "voice_intensity": 0.5,   // 0.0 (low) to 1.5 (extreme)
    "first_frame": {
      "setup_notes": "...",
      "reference": {
        "reference_characters": ["..."], // Visible characters
        "reference_backgrounds": ["..."],// Background behind speaker
        "reference_shots": ["..."],      // Prior shots for continuity
        "edit_prompt": "..."  // Static composition
      },
      "generation_prompt": null // Non-null only if no references
    },
    "video_prompt": "..."     // Camera motion and actions
  }]
}
\end{verbatim}
\vspace{-10pt}
\caption{Storyboard output structure produced by the scene director.}
\vspace{-10pt}
\label{lst:storyboard}
\end{figure}

\section{Script Inspection and Selection}
After all island iterations, scripts are inspected and a final evaluation is conducted by merging all islands into a single league. To select the top scripts for rendering, we run a round-robin tournament where each script competes against all others before a specialized critic committee tailored for an academic audience. Following this, the top four scripts are selected for production.

\section{Human Evaluation Protocol}
\label{sec:appendix_human_eval}
\begin{figure}[htbp]
\centering
\includegraphics[width=1.0\linewidth]{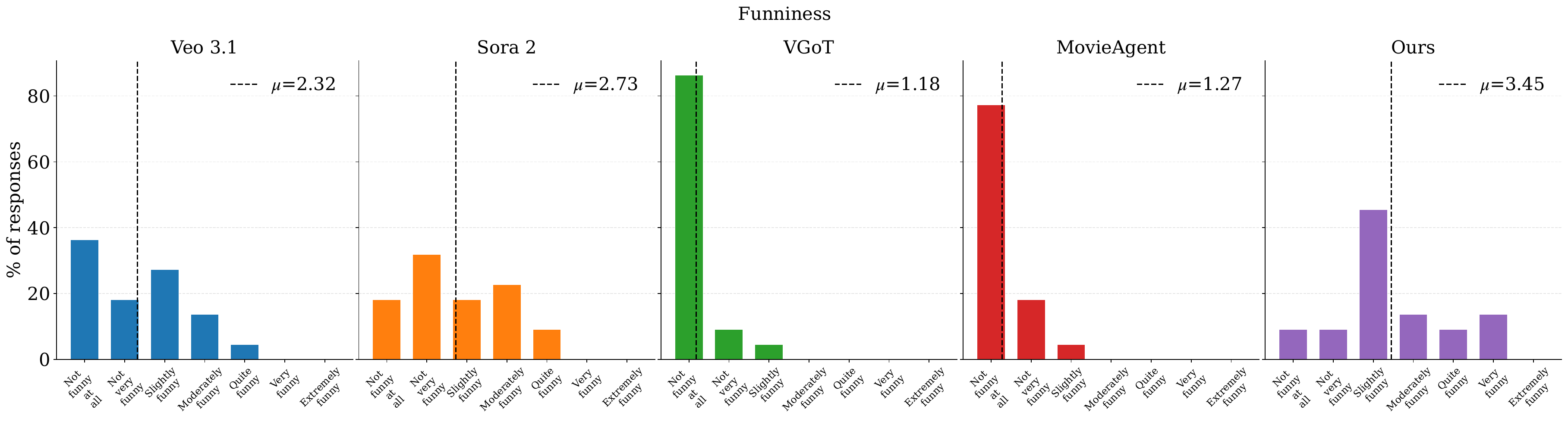}
\includegraphics[width=1.0\linewidth]{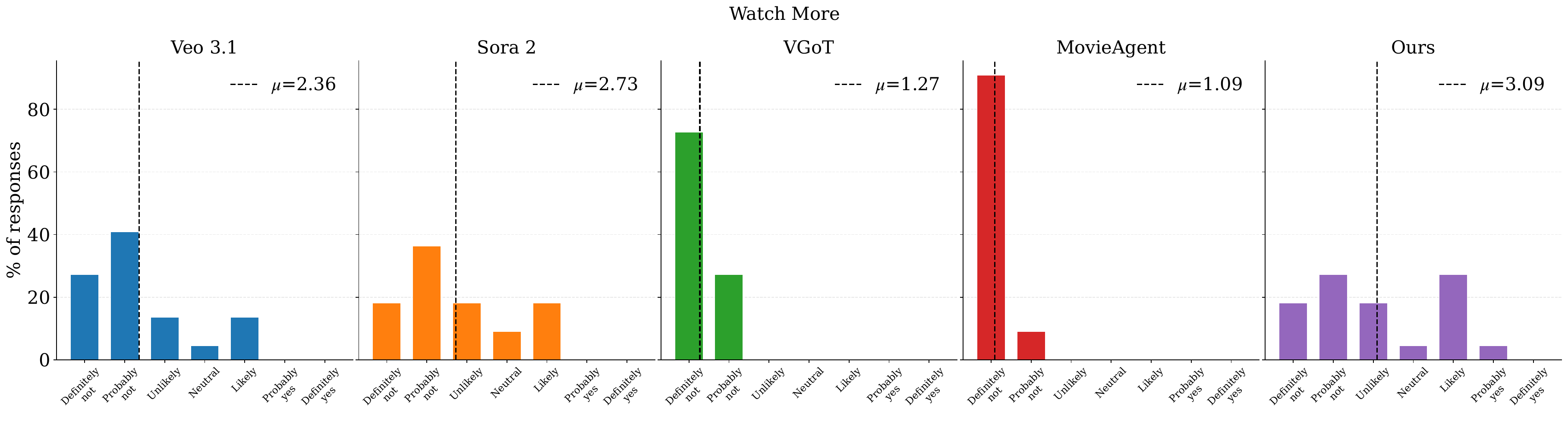}
\includegraphics[width=1.0\linewidth]{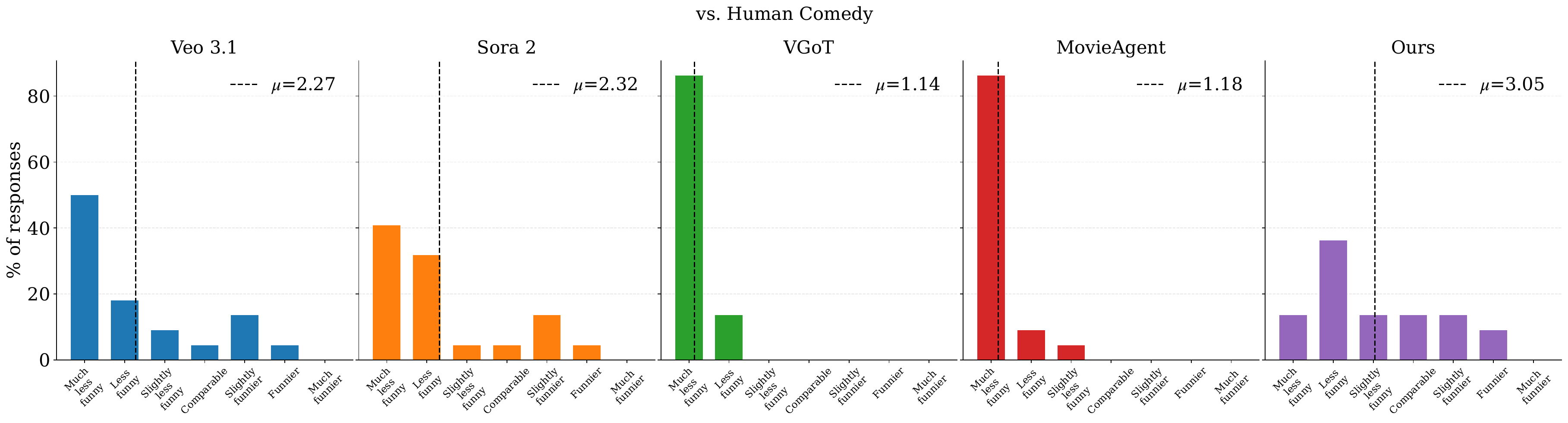}
\vspace{-20pt}
\caption{Distributions of human evaluation ratings for COMIC compared to baseline methods.}
\vspace{-10pt}
\label{fig:distribution}
\end{figure}

\paragraph{Baseline comparison.}
We evaluated five methods: COMIC (ours), VGoT~\citep{zheng2024videogen}, MovieAgent~\citep{wu2025automated}, Veo 3.1~\citep{google2025veo}, and Sora 2~\citep{openai2025sora}. We conducted a baseline video evaluation with participants across the US, Europe, and Asia, yielding $110 = 22 \times 5$ responses in total. Each method was represented by four videos, without manual curation. Figure~\ref{fig:distribution} illustrates the rating distributions, where we were able to obtain highly diverse opinions. Participants rated each video on a 7-point Likert scale across the following dimensions:
\begin{enumerate}
    \item \textbf{Funniness:} ``How funny did you find this video?''\\
    \textit{(1) Not funny at all $\to$ (2) Not very funny $\to$ (3) Slightly funny $\to$ (4) Moderately funny $\to$ (5) Quite funny $\to$ (6) Very funny $\to$ (7) Extremely funny}
    \item \textbf{Re-watch Intent:} ``Would you like to watch more videos like this?''\\
    \textit{(1) Definitely not $\to$ (2) Probably not $\to$ (3) Unlikely $\to$ (4) Neutral $\to$ (5) Likely $\to$ (6) Probably yes $\to$ (7) Definitely yes}
    \item \textbf{Comparison to Human Comedy:} ``Thinking of all the human-made sketch comedies you have ever seen, how funny is this video compared to the average human-made one?''\\
    \textit{(1) Much less funny $\to$ (2) Less funny $\to$ (3) Slightly less funny $\to$ (4) Comparable $\to$ (5) Slightly funnier $\to$ (6) Funnier $\to$ (7) Much funnier}
    \item \textbf{Script Quality:} ``How would you rate the funniness of the script?''\\
    \textit{(1) Not funny at all $\to$ (2) Slightly funny $\to$ (3) Somewhat funny $\to$ (4) Moderately funny $\to$ (5) Quite funny $\to$ (6) Very funny $\to$ (7) Extremely funny}
    \item \textbf{Narrative Quality:} ``How would you rate the narrative quality of the script (\eg, story arc, pacing)?''\\
    \textit{(1) Very poor $\to$ (2) Poor $\to$ (3) Slightly poor $\to$ (4) Neutral $\to$ (5) Slightly good $\to$ (6) Good $\to$ (7) Excellent}
    \item \textbf{Visual Realism:} ``How would you rate the visual realism of this video?''\\
    \textit{(1) Very poor $\to$ (2) Poor $\to$ (3) Slightly poor $\to$ (4) Neutral $\to$ (5) Slightly good $\to$ (6) Good $\to$ (7) Excellent}
    \item \textbf{Visual Consistency:} ``How would you rate the visual consistency of this video (\eg, characters, backgrounds)?''\\
    \textit{(1) Very poor $\to$ (2) Poor $\to$ (3) Slightly poor $\to$ (4) Neutral $\to$ (5) Slightly good $\to$ (6) Good $\to$ (7) Excellent}
\end{enumerate}

\paragraph{Ablation study.}
Participants completed a paired A/B comparison between COMIC and the critic-free baseline across five dimensions:
\begin{enumerate}
    \item \textbf{Funniness:} ``Which video do you find funnier?''
    \item \textbf{Script Quality:} ``Which video has a funnier script?''
    \item \textbf{Narrative Quality:} ``Which video has better narrative flow (\eg, story arc, pacing)?''
    \item \textbf{Visual Realism:} ``Which video has higher realism?''
    \item \textbf{Visual Consistency:} ``Which video has higher visual consistency (\eg, characters, backgrounds)?''
\end{enumerate}
Each question offered three choices: \textit{Video A}, \textit{About the same}, or \textit{Video B}. Neutral responses (\textit{About the same}) were excluded from the analysis.

\section*{AI Use Statement}
We used generative AI tools to polish the manuscript, refine grammar, format LaTeX, and assist with coding and implementation. We did not use these tools to develop the research framework, propose or refine research hypotheses, design experiments, process the collected dataset, or interpret results. The use of generative models within COMIC for content generation and evaluation is described in the methods and model specifications. The authors take responsibility for the final manuscript, implementation, and reported results.

\section*{Ethics Statement}
COMIC is intended for creative production of fictional sketch comedy. The characters in our generated videos are fictional. Nevertheless, systems that synthesize realistic audiovisual content could be misused for deepfakes, impersonation, or misleading representations. Generated content should be clearly identified as synthetic, and use of real individuals' likenesses or voices would require appropriate consent. Comedy generation can also reproduce stereotypes or produce offensive content; human review remains important before public distribution. Our intended use is fictional entertainment, and we do not endorse deceptive or harmful uses of the framework.

\section*{Reproducibility Statement}
We document the method, model choices, and hyperparameter configurations to support reproduction of our experiments. Table~\ref{tab:models} lists the models used at each stage, and Table~\ref{tab:scales} specifies the small, base, and large configurations. The appendix also describes writing-critic selection, the storyboard output structure, and the human evaluation protocol. These details represent our best effort to make the experimental setup reproducible; stochastic generation and changes to externally hosted models may affect exact outputs. We will release the full codebase upon acceptance.

\end{document}